\documentclass[lettersize,journal]{IEEEtran}
\usepackage{amsmath,amsfonts}
\usepackage{algorithm}
\usepackage{array}
\usepackage[caption=false,font=normalsize,labelfont=sf,textfont=sf]{subfig}
\usepackage{textcomp}
\usepackage{stfloats}
\usepackage{url}
\usepackage{verbatim}
\usepackage{graphicx}
\usepackage{cite}
\usepackage{graphicx}
\usepackage{graphicx}
\usepackage{booktabs}
\usepackage[table]{xcolor}
\usepackage{colortbl}
\usepackage{array}
\usepackage{algorithm}
\usepackage{algpseudocode}
\usepackage{mdframed}
 \usepackage{amsmath}
 \usepackage{subcaption}

\usepackage{threeparttable}

\definecolor{oursblue}{RGB}{230,243,250}
\definecolor{sectiongray}{RGB}{242,242,242}

\newcommand{\best}[1]{\textbf{#1}}
\newcommand{\second}[1]{\underline{#1}}

\begin{document}

\title{Drive-HWM: Hierarchical World Models for Dynamic-Latent Guided Autonomous Driving}

\author{
  Zhaoxin Fan,
  Tianbao Zhang,
  Wenjun Wu,
  Xiaofeng Wang,
  Yeying Jin,
  Jian Zhao,
  Zheng Zhu\thanks{
    Zhaoxin Fan and Wenjun Wu are with the School of Artificial Intelligence, Beihang University, Beijing, China.
    \par Tianbao Zhang is with Shanghai Jiao Tong University, Shanghai, China and Dim12 AI. 
    \par Xiaofeng Wang and Zheng Zhu are with GigaAI.
    \par Yeying Jin is with National University of Singapore.
    \par Jian Zhao is with TeleAI.
    \par Shuicheng Yan is with the National University of Singapore, Singapore.
    \par Jian Zhao and Zheng Zhu are the corresponding authors
    (Email: zhaoj90@chinatelecom.cn; zhengzhu@ieee.org).
  },
  Shuicheng Yan
}

% \author{IEEE Publication Technology,~\IEEEmembership{Staff,~IEEE,}
%         % <-this % stops a space
% \thanks{This paper was produced by the IEEE Publication Technology Group. They are in Piscataway, NJ.}% <-this % stops a space
% \thanks{Manuscript received April 19, 2021; revised August 16, 2021.}}

% % The paper headers
% \markboth{Journal of \LaTeX\ Class Files,~Vol.~14, No.~8, August~2021}%
% {Shell \MakeLowercase{\textit{et al.}}: A Sample Article Using IEEEtran.cls for IEEE Journals}

% \IEEEpubid{0000--0000/00\$00.00~\copyright~2021 IEEE}
% Remember, if you use this you must call \IEEEpubidadjcol in the second
% column for its text to clear the IEEEpubid mark.

\maketitle
\begin{figure*}[h]
\centering
\includegraphics[width=\linewidth]{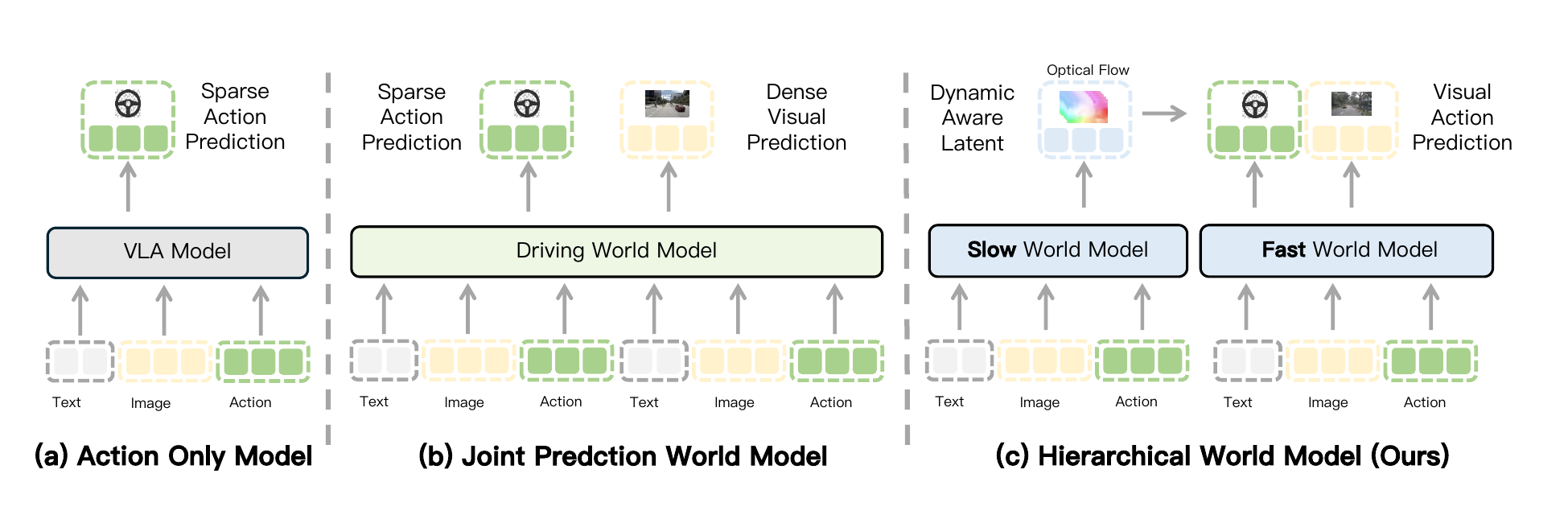}
\caption{
\textbf{Comparison of three world modeling paradigms for autonomous driving.}
(a) Action-only models rely on spare action prediction.
(b) Joint prediction world models jointly perform sparse action prediction and dense visual prediction.
(c) Our hierarchical world model separates slow and fast world modeling.
} \label{fig:overview}
\end{figure*}

\begin{abstract}
World models offer a promising paradigm for autonomous driving by predicting
how traffic scenes may evolve and using such predictions to support action
generation. However, existing approaches either separate future prediction
from action generation or jointly predict them at the same temporal scale,
making it difficult to simultaneously achieve long-horizon anticipation and
responsive, observation-grounded decision making. We present
\textbf{Drive-HWM}, a hierarchical slow--fast world modeling framework that
organizes future representation prediction and action generation at
complementary temporal scales. The slow world model predicts multi-step
future representations to capture extended scene evolution. To explicitly
model the abundant motion dynamics in driving environments, we introduce Dynamic-Aware Latents learned through optical-flow prediction. Guided
by these future representations, the fast model uses a lightweight multimodal
backbone and an autoregressive expert to jointly predict the next frame and
the immediate action from the latest observation. Next-frame prediction
encourages the fast model to capture imminent scene evolution, while
one-step action generation allows decisions to be continuously updated as
new observations arrive. Extensive experiments on NAVSIM v1 and v2
demonstrate the strong driving performance of Drive-HWM. Comprehensive
ablation studies further validate the effectiveness of the hierarchical
slow--fast design, dynamics-aware future representations, and joint
next-frame and action prediction. 

%Code is available at \url{https://github.com/your-account/Drive-HWM}.
\end{abstract}


\section{Introduction}
\IEEEPARstart{A}{utonomous} driving unfolds in the future, not in the present. Although an autonomous vehicle perceives only the current traffic scene, every decision it makes depends on what may happen next: whether a pedestrian will cross the road, whether a nearby vehicle will change lanes, and how the scene will respond to the vehicle's own actions. Reliable driving therefore requires more than recognizing the present; it requires reasoning about the evolution of the environment and the consequences of possible decisions. Yet, conventional driving systems largely operate reactively, mapping observed scenes directly to actions without explicitly modeling the future. World models provide a compelling alternative by learning the dynamics of driving environments and internally simulating how different futures may unfold~\cite{wote,univla,worldvla}. Such predictive reasoning enables an agent to evaluate candidate behaviors before acting, thereby bridging scene understanding and foresighted planning~\cite{drivedreamer,diffusiondrive,vad}. This capability makes world modeling a critical foundation for autonomous driving in complex, interactive, and uncertain environments.

In recent years, world modeling for autonomous driving has evolved from implicit future
reasoning to explicit future prediction. Broadly speaking, VLA-based driving
models~\cite{wote,worldvla} can be viewed as implicit world models: they
encode expectations about scene evolution in their latent representations,
but expose only driving actions or trajectories as outputs (Fig. \ref{fig:overview} (a)). Video- and
representation-based world models~\cite{drivedreamer,vista,gaia,driveworld}
make such expectations explicit by predicting future images or
representations, including latent features, occupancy states, and motion
fields. However, these models are primarily future predictors rather than
executable driving policies.

To support decision making, predictive world models are typically adapted
with action-generation modules or used indirectly as data generators,
learned simulators, or providers of reward and supervision signals. Future
prediction and action generation thus remain separated across model
components or training stages. Recent world-action models instead jointly
predict future states and driving actions within a unified
architecture~\cite{uniad}, encouraging the learned dynamics to capture
decision-relevant scene evolution while providing richer temporal
supervision for action generation (Fig. \ref{fig:overview} (b). Nevertheless, their benefits are often
limited to short horizons. As the horizon extends, prediction uncertainty
and rollout errors accumulate, progressively degrading future
representations and their utility for decision making. Joint prediction
alone therefore cannot fully reconcile long-horizon anticipation with
immediate, observation-grounded control.

A key question then arises: how should future representation prediction and
action prediction be organized within a driving world model? Although closely
related, they serve distinct roles at different temporal scales. Future
representation prediction aims to capture how a driving scene evolves over
an extended horizon. Since such evolution is largely governed by the motion
of the ego vehicle and surrounding agents, as well as their interactions,
the predicted representations should preserve not only scene semantics but
also temporal dynamics. Multi-step prediction can thereby describe a
coherent future evolution, such as the progression of a complete left turn.
Action prediction, by contrast, is local and high-frequency: each action
should be grounded in the latest observation to respond promptly to changing
conditions. Predicting a long action sequence at once may accumulate
execution errors and gradually deviate from the intended behavior. Future
representation prediction and action prediction should therefore remain
temporally distinct yet tightly coupled, with multi-step dynamics providing
predictive guidance for observation-grounded, one-step action generation.

To this end, we present \textbf{Drive-HWM}, a hierarchical slow--fast world
modeling framework that couples future representation prediction with
high-frequency action generation. The slow world model predicts multi-step
future representations to characterize scene evolution over an extended
horizon. To make these representations sensitive to motion, we introduce
\emph{Dynamic-Aware Latents}, which are learned through optical-flow
prediction to explicitly encode the dynamics of the ego vehicle and
surrounding agents. Conditioned on these future representations and the
latest observation, the fast model employs a lightweight multimodal backbone
and a fast autoregressive expert to jointly predict the next frame and the
immediate action. Next-frame prediction grounds the fast model in the
imminent evolution of the scene, promoting consistency between its local
world understanding and action generation. Meanwhile, predicting only the
immediate action allows the model to revise its decision whenever a new
observation arrives. Through this hierarchical coupling, Drive-HWM combines
multi-step dynamics-aware future prediction with responsive,
observation-grounded action generation.

To evaluate Drive-HWM, we conduct extensive experiments on  NAVSIM v1/v2. Drive-HWM achieves strong performance across
multiple driving metrics, demonstrating the effectiveness of hierarchical
slow--fast world modeling. Comprehensive ablation studies further validate
the contributions of dynamics-aware future representation prediction, joint
next-frame and action prediction, and their hierarchical coupling. Our main
contributions are summarized as follows:
\begin{itemize}
    \item We propose Drive-HWM, a hierarchical slow--fast world modeling
    framework that couples multi-step future representation prediction with
    high-frequency, observation-grounded action generation.

    \item We develop dynamics-aware future representation prediction for the
    slow world model and joint next-frame and immediate-action prediction for
    the fast model, enabling long-horizon anticipation and responsive action
    generation at complementary temporal scales.

    \item Extensive experiments demonstrate the strong driving performance
    of Drive-HWM, while comprehensive ablations validate its key architectural
    and supervisory designs.
\end{itemize}

\section{Related Work}
In this section We review three lines of research most relevant to our work:: vision--language--action models for autonomous driving, predictive world models for autonomous driving, and world--action models.

\textbf{Vision--Language--Action Models for Autonomous Driving.}
Vision--language--action (VLA) models extend vision--language models from
scene understanding and semantic reasoning to executable driving decisions.
Early approaches used language models to interpret traffic scenes, explain
driving behaviors, or suggest high-level maneuvers without directly producing
executable actions~\cite{11,12,13}. Subsequent methods connected semantic
reasoning to low-level control through modular pipelines, whose discrete or
non-differentiable interfaces limited end-to-end optimization
~\cite{14,15,16,17}. Recent end-to-end VLAs instead directly map multimodal
observations and instructions to trajectories or control commands
~\cite{18,19,20,21,drivemoe,diffvla,recogdrive,autovla}. For example,
DriveMoE~\cite{drivemoe} employs scene- and skill-specialized experts,
ReCogDrive~\cite{recogdrive} combines vision--language reasoning with a
diffusion planner, and AutoVLA~\cite{autovla} unifies reasoning and action
generation through autoregressive action tokens. Broadly, VLAs may be viewed as implicit world models, as action generation
requires representations that capture scene dynamics and potential action
consequences. Several driving VLAs further adopt fast--slow designs to balance
decision quality and computational cost: routine scenarios use direct action
generation, whereas challenging situations invoke more expensive semantic or
chain-of-thought reasoning
~\cite{luo2025adathinkdrive,autovla,qian2024fasionad}. These methods separate
reasoning modes and allocate computation according to scenario complexity.
In contrast, our hierarchy separates explicit future representation
prediction from action generation according to their temporal roles: the slow
model anticipates extended scene evolution, while the fast model produces
observation-grounded actions. In the following, we therefore use
\emph{world model} in a narrower sense to denote models that explicitly
predict future observations, states, or representations.

\textbf{Predictive World Models for Autonomous Driving.}
World models learn environment dynamics by predicting future observations or
latent states from historical context, optionally conditioned on agent
actions~\cite{awais2025foundation, lin2025navcot,wang2024jarvis}. In autonomous
driving, existing methods differ primarily in the representation space in
which the future is predicted. Generative world models such as
GAIA-1~\cite{gaia}, DriveDreamer~\cite{drivedreamer}, and
Vista~\cite{vista} synthesize future camera observations, providing
realistic and controllable simulations for data generation and policy
evaluation. Other approaches predict more compact representations.
Copilot4D~\cite{col4d} models future visual tokens, while occupancy-based
world models forecast future 3D occupancy states to capture spatial scene
evolution. DriveWorld~\cite{driveworld} learns future latent dynamics as a
self-supervised objective for transferable spatiotemporal representation
learning. These methods demonstrate that predicting future observations or
representations provides rich supervision for learning scene dynamics.
Nevertheless, their primary output is the future world state rather than an
executable driving action. Consequently, depending on their design goals,
they can serve as data generators~\cite{zeng2026rethinking}, learned
simulators~\cite{zhang2026world}, representation pretraining
objectives~\cite{he2026pre}, or auxiliary modules for downstream
planning~\cite{jiang2025irl}. Moreover, although some methods condition future
prediction on a given action or trajectory, the action serves as an input
describing how the future should unfold rather than a prediction target
jointly generated by the world model. Our work builds on future
representation prediction but focuses on how its extended temporal context
can be coupled with high-frequency action generation.

\textbf{World--Action Models.}
World--action models bridge predictive world modeling and policy learning by
jointly modeling future states and agent actions \cite{ye2026gigaworld,bi2026motus,zhu2026wmpo}. In robotics,
UniVLA~\cite{univla} autoregressively models vision, language, and actions in
a shared token space and incorporates future prediction to learn causal
environment dynamics. WorldVLA~\cite{worldvla} similarly unifies future-image
and action generation, allowing world prediction and policy learning to
provide mutual supervision. In autonomous driving, LAW~\cite{law} predicts
future latent scene features conditioned on the current representation and
predicted ego trajectory, using future observations to jointly improve
representation learning and trajectory prediction. Other methods couple
predicted future states with candidate action evaluation or directly
generate future videos and driving actions within a unified
model~\cite{li2025end,wang2025prophetdwm}. By connecting state evolution with action generation, these methods learn
more decision-relevant dynamics than standalone future predictors. However,
existing approaches generally rely on a single predictive architecture or
model future states and actions within a common rollout process. This
organization overlooks their different temporal requirements: future
representations should capture extended scene evolution, whereas executable
actions must respond promptly to the latest observation. Long autoregressive
action rollouts may also propagate early prediction errors to subsequent
decisions. Drive-HWM addresses this distinction through a hierarchical
slow--fast architecture: the slow world model predicts multi-step,
dynamics-aware future representations, while the lightweight fast model
jointly predicts the next frame and only the immediate action. This design
uses long-horizon anticipation to guide action generation without sacrificing
high-frequency observation grounding.

\begin{figure*}[t]
\centering
\includegraphics[width=\linewidth]{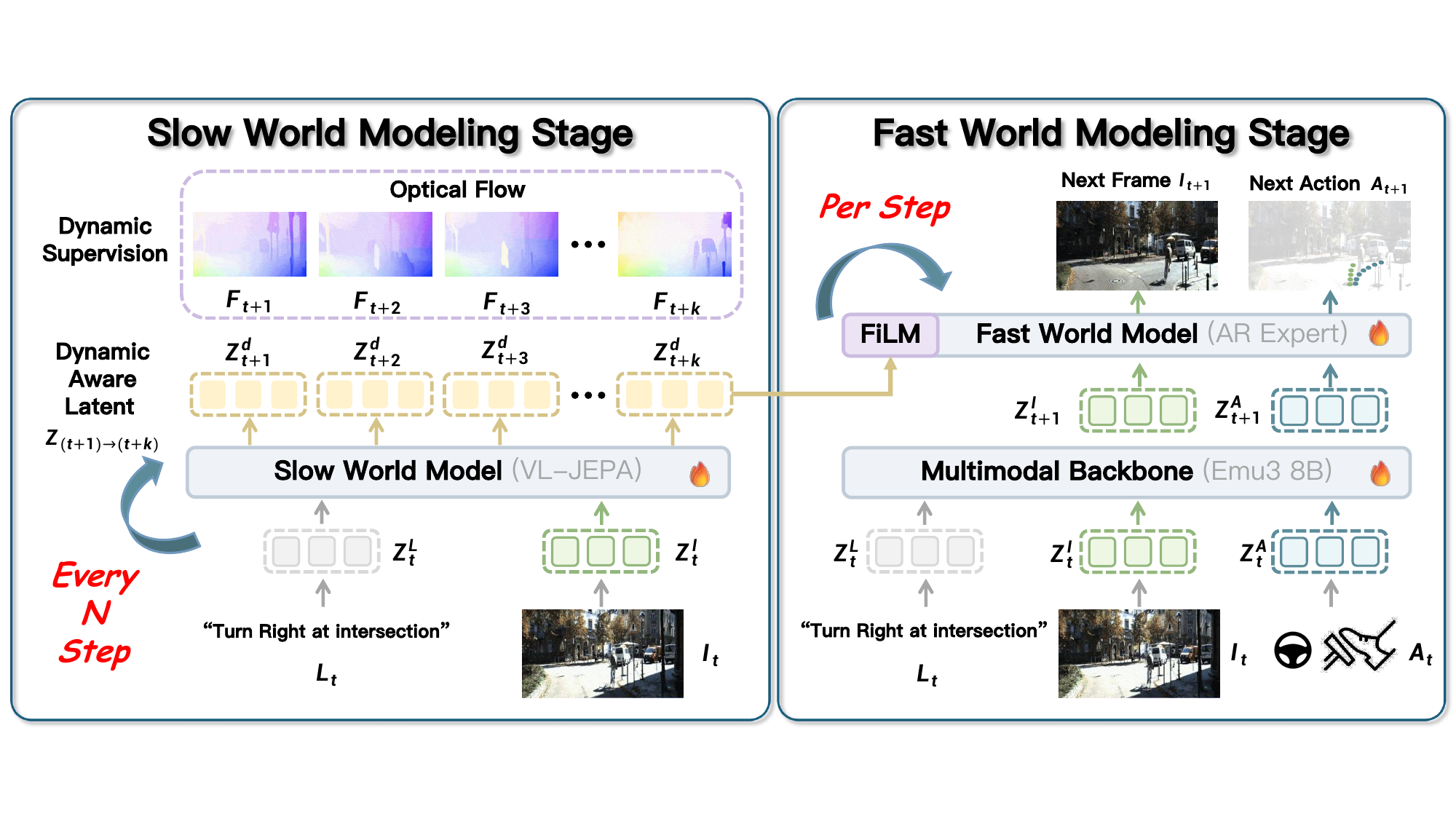}
\vspace{-0.5in}
\caption{\textbf{Overview of the proposed slow--fast world modeling framework.}
The slow world model, updated every $N$ steps, employs VL-JEPA to predict multi-horizon optical flow under dynamic supervision, producing Dynamic-Aware Latents $\mathbf{Z}^{d}_{(t+1)\rightarrow(t+k)}$ that encode anticipated scene evolution. At every step, the temporally aligned dynamic latent is injected via FiLM into the fast world model built upon Emu3-8B. The autoregressive expert integrates the current instruction, observation, and action history to jointly predict the next action and next-frame visual tokens, enabling long-horizon dynamic awareness while remaining responsive to the latest observations.}\label{main}
% \caption{The architecture of Drive-HWM, a hierarchical slow--fast world modeling framework for autonomous driving. 
% The system consists of two interacting components: (a) a Slow Planning World Model that performs dynamic-aware planning by predicting future motion representations under optical-flow supervision, and (b) a Fast Controller that generates observation-grounded actions at each timestep by conditioning on both the current observation and the planned latent states.} \label{main}
\end{figure*}

\section{Method}
\label{sec:method}

This section presents Drive-HWM. We first formulate the driving problem
and provide an overview of the proposed slow--fast hierarchy. We then
describe the slow world model, which anticipates extended scene evolution
in latent space, and the fast world model, which generates actions grounded
in both the latest observation and the predicted future context. Finally,
we introduce the training objectives used to optimize the two components.

\subsection{Problem Formulation and Framework Overview}
\label{sec:overview}

We formulate autonomous driving as sequential decision-making under partial
observability. At timestep $t$, the agent receives an observation
$o_t \in \mathcal{O}$, which may contain multi-view images and auxiliary
sensory inputs, and predicts an action $a_t \in \mathcal{A}$. Since a single
observation does not fully characterize the underlying environment state,
the decision is conditioned on the interaction history
\begin{equation}
    \mathcal{H}_t
    =
    \left(o_{\leq t},a_{<t}\right),
    \qquad
    h_t = E_{\eta}(\mathcal{H}_t),
\end{equation}
where $E_{\eta}$ encodes the available observations and previous actions into
a contextual representation $h_t$. The objective is to learn a policy that
predicts driving actions while accounting for how the observed scene may
evolve over time.

A single-rate model must use the same representation and computational
schedule for two different requirements: capturing extended scene evolution
and responding to instantaneous observations. Drive-HWM addresses this
mismatch through two world models operating at different temporal scales.
The \emph{slow world model} is evaluated once every $N$ timesteps and predicts
a sequence of future latent representations over a horizon of $K$ steps.
The \emph{fast world model} runs at every timestep and generates an action
conditioned on the latest observation history and a temporally aligned latent
prediction from the slow model. 

%In our implementation, $K=N=8$.

More formally, let
\begin{equation}
    \tau(t) = N\left\lfloor \frac{t}{N} \right\rfloor
\end{equation}
denote the most recent slow-model update preceding timestep $t$. At update
time $\tau$, the slow model predicts
$\hat{Z}^{s}_{\tau}
=\{\hat{z}^{s}_{\tau+k\mid\tau}\}_{k=1}^{K}$.
The prediction aligned with the next state at timestep $t$ is therefore
\begin{equation}
    q_t^{s}
    =
    \hat{z}^{s}_{t+1\mid\tau(t)}.
\end{equation}
Drive-HWM can then be summarized by the following hierarchical factorization:
\begin{equation}
\begin{aligned}
&
p_{\Theta}\!\left(
    \{a_t,z^{f}_{t+1}\}_{t=0}^{T-1},
    \{Z^{s}_{\tau}\}_{\tau\in\mathcal{R}}
    \,\middle|\,
    o_{\leq T}
\right)
\\
&\quad =
\prod_{\tau\in\mathcal{R}}
p_{\theta_s}\!\left(
    Z^{s}_{\tau}
    \mid
    \mathcal{H}_{\tau}
\right)
\prod_{t=0}^{T-1}
p_{\theta_f}\!\left(
    a_t,z^{f}_{t+1}
    \mid
    \mathcal{H}_t,q_t^{s}
\right),
\end{aligned}
\label{eq:hierarchical_factorization}
\end{equation}
where $\mathcal{R}=\{0,N,2N,\ldots\}$ is the set of slow-model update
timesteps. The slow prediction is reused within each update interval, whereas
the fast model continuously incorporates newly observed evidence. Consequently,
the predicted future provides a persistent long-range context without
preventing the action model from correcting its decisions when unexpected
scene changes occur. The framework of Drive-HWM is illustrated in Fig. \ref{main}. Next, we introduce the slow and fast wolrd model indetail.

\subsection{Slow World Model: Dynamic-Aware Future Prediction}
\label{sec:slow_world_model}

The slow world model is designed to capture long-term scene dynamics rather
than appearance variations. A straightforward solution would be to predict
future visual features extracted from RGB observations. However, such
features are typically dominated by appearance semantics and spatial content,
while the information most relevant to driving---including ego-motion,
object displacement, and their temporal evolution---may occupy only a small
portion of the representation. Moreover, directly predicting future RGB
frames requires modeling substantial appearance details that are difficult
to extrapolate and are not necessarily useful for action generation.

To address this issue, we introduce a \emph{Dynamic-Aware Latent} (DAL)
representation learned by predicting future optical flow. Optical flow
provides an explicit description of pixel-wise displacement and naturally
emphasizes moving agents, ego-motion-induced scene changes, and the geometric
evolution of the observed environment. Compared with future RGB prediction,
it suppresses static appearance details such as texture, illumination, and
color, allowing the slow model to allocate its capacity to temporally
predictive motion information.

\paragraph{Multi-step optical-flow prediction.}
Let $\tau$ denote a slow-model update timestep and let
\begin{equation}
    h_\tau = E_\eta(\mathcal{H}_\tau),
    \qquad
    \mathcal{H}_\tau=(o_{\leq \tau},a_{<\tau}),
\end{equation}
be the contextual representation of the available interaction history. Given
$h_\tau$, the slow world model predicts scene dynamics over the subsequent
$K$ temporal offsets:
\begin{equation}
    \left\{
        \hat{d}^{s}_{\tau+k\mid\tau},
        \hat{\mathcal{F}}_{\tau+k\mid\tau}
    \right\}_{k=1}^{K}
    =
    S_{\theta_s}(h_\tau),
\label{eq:slow_flow_prediction}
\end{equation}
where $\hat{d}^{s}_{\tau+k\mid\tau}$ denotes the Dynamic-Aware Latent at
future offset $k$, and
$\hat{\mathcal{F}}_{\tau+k\mid\tau}\in\mathbb{R}^{H_f\times W_f\times 2}$
is the corresponding predicted optical-flow field. The two channels of
$\hat{\mathcal{F}}_{\tau+k\mid\tau}$ represent horizontal and vertical
displacements, respectively.

For each temporal offset, the flow prediction is decoded from the
corresponding Dynamic-Aware Latent:
\begin{equation}
    \hat{\mathcal{F}}_{\tau+k\mid\tau}
    =
    D_{\mathrm{flow}}
    \left(
        \hat{d}^{s}_{\tau+k\mid\tau}
    \right),
    \qquad k=1,\ldots,K,
\label{eq:flow_decoder}
\end{equation}
where $D_{\mathrm{flow}}$ is a lightweight flow decoder. The intermediate
representation $\hat{d}^{s}_{\tau+k\mid\tau}$ is therefore not an
unconstrained visual feature. It is explicitly optimized to retain the
information required to recover future motion fields and is consequently
aware of both spatial content and temporal displacement.

The supervision target associated with future offset $k$ is defined as
\begin{equation}
    \mathcal{F}_{\tau+k}
    =
    \operatorname{Flow}
    \left(
        o_{\tau+k-1},
        o_{\tau+k}
    \right),
\label{eq:flow_target}
\end{equation}
where $\operatorname{Flow}(\cdot,\cdot)$ denotes the optical-flow estimator
used to construct the training targets. All $K$ future flow fields are
predicted from the same observed history $h_\tau$, rather than being generated
by recursively feeding previously predicted flow into the model:
\begin{equation}
    p_{\theta_s}
    \left(
        \mathcal{F}_{\tau+1:\tau+K}
        \mid
        \mathcal{H}_\tau
    \right)
    =
    \prod_{k=1}^{K}
    p_{\theta_s}
    \left(
        \mathcal{F}_{\tau+k}
        \mid
        \mathcal{H}_\tau,k
    \right).
\label{eq:slow_factorization}
\end{equation}
This parallel multi-offset formulation avoids recursive flow warping and
prevents prediction errors at early offsets from being directly propagated
to later ones.

\paragraph{Dynamic-Aware Latent.}
Although dense optical flow provides direct motion supervision, passing raw
flow fields to the fast model would discard contextual cues needed to
interpret the motion. For example, similar image displacement may correspond
to different driving implications depending on whether it originates from a
vehicle, a pedestrian, the road surface, or camera motion. Drive-HWM therefore
uses the hidden representation before flow decoding as its interface between
the two temporal levels:
\begin{equation}
    D_\tau^s
    =
    \left\{
        \hat{d}^{s}_{\tau+1\mid\tau},
        \ldots,
        \hat{d}^{s}_{\tau+K\mid\tau}
    \right\}.
\label{eq:dal_sequence}
\end{equation}

Because $D_\tau^s$ must support the reconstruction of multi-step future flow,
it encodes anticipated scene displacement while preserving the contextual
information required to explain that displacement. We refer to
$D_\tau^s$ as the \emph{Dynamic-Aware Latent sequence}. In contrast to generic
RGB features, which primarily describe what is currently visible,
$D_\tau^s$ emphasizes how the visible scene is likely to change over time.

At a fast-model timestep $t$, we select the temporally aligned Dynamic-Aware
Latent from the most recent slow update:
\begin{equation}
    \tau(t)
    =
    N\left\lfloor\frac{t}{N}\right\rfloor,
    \qquad
    q_t^s
    =
    \hat{d}^{s}_{t+1\mid\tau(t)}.
\label{eq:dal_alignment}
\end{equation}
The aligned latent $q_t^s$, rather than the decoded optical-flow field, is
provided to the fast world model. In this manner, optical flow acts as an
explicit learning signal for discovering dynamics-sensitive
representations, while the latent interface retains richer information for
subsequent action generation.

\paragraph{Flow prediction objective.}
The slow world model is optimized by matching its multi-step predictions to
the corresponding future optical-flow targets:
\begin{equation}
    \mathcal{L}_{\mathrm{flow}}
    =
    \frac{1}{K}
    \sum_{k=1}^{K}
    w_k\,
    \ell_{\mathrm{flow}}
    \left(
        \hat{\mathcal{F}}_{\tau+k\mid\tau},
        \mathcal{F}_{\tau+k}
    \right),
\label{eq:flow_prediction_loss}
\end{equation}
where $w_k$ controls the contribution of different prediction offsets and
$\ell_{\mathrm{flow}}$ denotes the optical-flow prediction loss. When flow
validity or visibility masks are available, the loss is evaluated only over
valid locations:
\begin{equation}
    \ell_{\mathrm{flow}}
    \left(
        \hat{\mathcal{F}},
        \mathcal{F}
    \right)
    =
    \frac{
        \sum_{u}
        M(u)\,
        \rho\!\left(
            \hat{\mathcal{F}}(u)-\mathcal{F}(u)
        \right)
    }{
        \sum_{u} M(u)+\epsilon
    },
\label{eq:masked_flow_loss}
\end{equation}
where $u$ indexes spatial locations, $M$ is a valid-flow mask, and
$\rho(\cdot)$ is the employed robust regression penalty.

The flow prediction objective serves two complementary purposes. First, it
provides explicit supervision for learning long-range scene dynamics without
requiring photorealistic future reconstruction. Second, it shapes the
intermediate latent space into a dynamics-sensitive representation that can
be reused by the fast world model. The slow branch therefore does not output
actions or an explicit trajectory. Instead, it supplies a sequence of
future-oriented Dynamic-Aware Latents that informs high-frequency action
generation.

\subsection{Fast World Model: Dynamics-Conditioned Action Prediction}
\label{sec:fast_world_model}

The slow world model predicts how the driving scene may evolve over an
extended temporal horizon. However, its prediction is computed at a lower
frequency and cannot incorporate observations arriving after the latest slow
update. The fast world model therefore operates at every timestep to generate
actions from the most recent visual evidence, while using the
Dynamic-Aware Latent (DAL) supplied by the slow model as predictive context.

We instantiate the fast world model using Emu3-8B~\cite{emu3} as the
action-prediction backbone. Emu3 represents visual content as discrete tokens
and models multimodal sequences through autoregressive next-token prediction.
This formulation provides a unified interface for processing the current
observation and predicting both action and future visual tokens. On top of the
Emu3 backbone, we introduce FiLM-based dynamic conditioning and an
autoregressive expert for driving-specific prediction.

\paragraph{Visual token representation.}
At timestep $t$, the latest observation $o_t$ is converted into a sequence of
discrete visual tokens using the Emu3 visual tokenizer:
\begin{equation}
    V_t
    =
    \mathcal{T}_{\mathrm{vis}}(o_t)
    =
    \left(
        v_{t,1},\ldots,v_{t,L_v}
    \right),
\label{eq:fast_visual_tokens}
\end{equation}
where $L_v$ denotes the number of visual tokens. Together with the preceding
observation and action context, these tokens are processed by the Emu3
backbone:
\begin{equation}
    H_t
    =
    B_{\theta_b}
    \left(
        V_{\leq t}, A_{<t}
    \right),
\label{eq:emu3_backbone}
\end{equation}
where $B_{\theta_b}$ denotes the Emu3-8B backbone, $A_{<t}$ contains the
previous action tokens, and
$H_t\in\mathbb{R}^{L_t\times D}$ represents the resulting contextual hidden
states. In contrast to using a separately designed visual encoder, this
token-based representation preserves the native visual modeling capability
of the pretrained multimodal backbone.

\paragraph{FiLM-based dynamic conditioning.}
The visual representation $H_t$ describes the most recently observed scene,
whereas the slow world model provides an anticipation of its future dynamics.
Let
\begin{equation}
    \tau(t)
    =
    N\left\lfloor\frac{t}{N}\right\rfloor,
    \qquad
    q_t^s
    =
    \hat{d}^{s}_{t+1\mid\tau(t)}
\label{eq:fast_dal_alignment}
\end{equation}
denote the Dynamic-Aware Latent aligned with the next transition at timestep
$t$. Rather than directly concatenating $q_t^s$ with the input sequence, we
inject it into the Emu3 hidden representation through Feature-wise Linear
Modulation (FiLM). Specifically, the modulation parameters are predicted from
the aligned DAL:
\begin{equation}
    \left(
        \gamma_t^{(l)},\beta_t^{(l)}
    \right)
    =
    G_{\mathrm{FiLM}}^{(l)}
    \left(
        q_t^s
    \right),
\label{eq:film_parameters}
\end{equation}
where $G_{\mathrm{FiLM}}^{(l)}$ is a lightweight projection module associated
with the $l$-th modulated transformer layer. The corresponding hidden states
are modulated as
\begin{equation}
    \widetilde{H}_t^{(l)}
    =
    \left(
        1+\gamma_t^{(l)}
    \right)
    \odot
    \operatorname{LN}
    \left(
        H_t^{(l)}
    \right)
    +
    \beta_t^{(l)},
\label{eq:film_modulation}
\end{equation}
where $\odot$ denotes element-wise multiplication and
$\operatorname{LN}(\cdot)$ denotes layer normalization.

This design allows the predicted dynamics to adaptively rescale and shift the
feature channels used for action prediction without changing the token
organization of the pretrained backbone. The current observation remains the
primary input to the fast model, while the DAL modulates its interpretation
according to the scene evolution anticipated by the slow world model. As a
result, the fast model combines up-to-date visual evidence with future-oriented
dynamic information instead of treating the slow prediction as an explicit
action command.

\paragraph{Autoregressive action expert.}
The FiLM-modulated features are subsequently processed by a driving-specific
autoregressive expert:
\begin{equation}
    R_t
    =
    E_{\theta_e}^{\mathrm{AR}}
    \left(
        \widetilde{H}_t
    \right),
\label{eq:ar_expert}
\end{equation}
where $E_{\theta_e}^{\mathrm{AR}}$ denotes the AR expert and $R_t$ is the
driving-specific representation used for output prediction. We represent the
action at timestep $t$ as an ordered token sequence
\begin{equation}
    A_t
    =
    \left(
        a_{t,1},\ldots,a_{t,L_a}
    \right),
\end{equation}
where the tokens encode the executable driving output. The action probability
is factorized autoregressively as
\begin{equation}
    p_{\theta_f}
    \left(
        A_t
        \mid
        \mathcal{H}_t,q_t^s
    \right)
    =
    \prod_{j=1}^{L_a}
    p_{\theta_f}
    \left(
        a_{t,j}
        \mid
        a_{t,<j},
        R_t
    \right).
\label{eq:action_autoregression}
\end{equation}
Accordingly, the action prediction objective is
\begin{equation}
    \mathcal{L}_{\mathrm{act}}
    =
    -\frac{1}{L_a}
    \sum_{j=1}^{L_a}
    \log
    p_{\theta_f}
    \left(
        a_{t,j}
        \mid
        a_{t,<j},
        R_t
    \right).
\label{eq:action_loss}
\end{equation}

The AR expert specializes the general-purpose multimodal representation of
Emu3 for driving action generation. Autoregressive factorization also models
the dependencies among action tokens, rather than estimating each action
component independently.

\paragraph{Next-frame visual supervision.}
Action supervision alone only indicates which control output should be
selected, providing limited information about the scene transition underlying
that decision. In particular, an action-only model may learn a direct mapping
from frequently occurring visual patterns to driving commands without
explicitly retaining how surrounding agents and scene geometry are changing.
To provide a denser learning signal for short-term dynamics, we additionally
require the fast model to predict the next observed frame.

The next observation $o_{t+1}$ is converted into target visual tokens using
the same visual tokenizer:
\begin{equation}
    V_{t+1}
    =
    \mathcal{T}_{\mathrm{vis}}(o_{t+1})
    =
    \left(
        v_{t+1,1},\ldots,v_{t+1,L_v}
    \right).
\label{eq:next_frame_tokens}
\end{equation}
The AR expert predicts these tokens conditioned on the current visual
evidence, the Dynamic-Aware Latent, and the action at the current timestep:
\begin{equation}
\begin{aligned}
&
p_{\theta_f}
\left(
    V_{t+1}
    \mid
    \mathcal{H}_t,q_t^s,A_t
\right)
\\
&\qquad =
\prod_{i=1}^{L_v}
p_{\theta_f}
\left(
    v_{t+1,i}
    \mid
    v_{t+1,<i},
    A_t,
    R_t
\right).
\end{aligned}
\label{eq:next_frame_autoregression}
\end{equation}
The corresponding visual prediction loss is
\begin{equation}
    \mathcal{L}_{\mathrm{img}}
    =
    -\frac{1}{L_v}
    \sum_{i=1}^{L_v}
    \log
    p_{\theta_f}
    \left(
        v_{t+1,i}
        \mid
        v_{t+1,<i},
        A_t,
        R_t
    \right).
\label{eq:next_frame_loss}
\end{equation}

This auxiliary objective serves three purposes. First, it requires the shared
representation to retain sufficient information to explain the immediate
scene transition, thereby complementing the long-horizon motion information
learned by the slow world model. Second, it provides spatially dense
supervision beyond the low-dimensional action labels, regularizing the action
expert against shortcut learning. Third, because the future visual prediction
is conditioned on the DAL, it encourages the fast model to make effective use
of the dynamic context instead of ignoring it during action training.

Importantly, next-frame prediction is introduced as an auxiliary training
objective rather than an additional output required by the driving system.
During inference, only the autoregressively predicted action tokens need to
be decoded, and future image generation can be omitted.

\paragraph{Fast-world-model objective.}
The fast world model is jointly optimized for action generation and
next-frame visual prediction:
\begin{equation}
    \mathcal{L}_{\mathrm{fast}}
    =
    \mathcal{L}_{\mathrm{act}}
    +
    \lambda_{\mathrm{img}}
    \mathcal{L}_{\mathrm{img}},
\label{eq:fast_world_model_loss}
\end{equation}
where $\lambda_{\mathrm{img}}$ controls the contribution of the auxiliary
visual prediction task. The action objective teaches the model which action
to execute, while the visual objective encourages it to capture the
short-term scene transition associated with that decision. Together with the
long-horizon DAL supplied by the slow world model, the fast branch combines
future-oriented dynamic context, current visual evidence, and high-frequency
action generation within a unified autoregressive framework.

\subsection{Training Objective}
\label{sec:training}

Drive-HWM is trained using supervision for long-horizon latent prediction,
short-horizon latent prediction, and action generation. For the fast world
model, we define
\begin{equation}
    \mathcal{L}_{\mathrm{fast}}
    =
    \mathcal{L}_{\mathrm{act}}
    \left(
        \hat{a}_t,a_t
    \right)
    +
    \lambda_f
    d\!\left(
        \hat{z}^{f}_{t+1},
        y_{t+1}
    \right),
\label{eq:fast_loss}
\end{equation}
where $\mathcal{L}_{\mathrm{act}}$ denotes the action prediction loss and
$\lambda_f$ balances action supervision against one-step latent prediction.
For autoregressively tokenized actions, $\mathcal{L}_{\mathrm{act}}$ is the
token-level negative log-likelihood; it can equivalently be instantiated as
a regression loss when continuous actions are directly predicted.

The complete objective is
\begin{equation}
    \mathcal{L}
    =
    \lambda_s\mathcal{L}_{\mathrm{slow}}
    +
    \mathcal{L}_{\mathrm{act}}
    +
    \lambda_f\mathcal{L}_{\mathrm{vis}},
\label{eq:total_loss}
\end{equation}
where
$\mathcal{L}_{\mathrm{vis}}
=d(\hat{z}^{f}_{t+1},y_{t+1})$
and $\lambda_s$ and $\lambda_f$ control the two predictive objectives.

During training, each slow prediction is shared by the $N$ fast-model steps
within the corresponding update interval, matching the temporal schedule
used at inference. The fast model is conditioned on the \emph{predicted}
slow latent rather than the encoded ground-truth future representation,
avoiding a discrepancy between training and deployment. At inference time,
the slow world model is evaluated only at timesteps in $\mathcal{R}$, while
the cached latent sequence is temporally indexed and supplied to the fast
world model at every intermediate timestep. This design amortizes
long-horizon future prediction across multiple action steps while retaining
observation-grounded action generation at the native control frequency.

% ==================== Required packages ====================

\begin{table*}[ht]
    \centering
    \caption{\textbf{Comparison with state-of-the-art methods on NAVSIM v1.}
    The best and second-best results are highlighted in bold and underlined,
    respectively. Our method is highlighted in light blue.}
    \label{tab:navsim_v1_sota}

    \small
    \setlength{\tabcolsep}{5.2pt}
    \renewcommand{\arraystretch}{1.12}

    \begin{tabular}{@{}lccrrrrrr@{}}
        \toprule
        \textbf{Method}
        & \textbf{Ref.}
        & \textbf{Sensors}
        & \textbf{NC$\uparrow$}
        & \textbf{DAC$\uparrow$}
        & \textbf{TTC$\uparrow$}
        & \textbf{C.$\uparrow$}
        & \textbf{EP$\uparrow$}
        & \textbf{PDMS$\uparrow$} \\
        \midrule

        Human
        & --
        & --
        & 100.0
        & 100.0
        & 100.0
        & 99.9
        & 87.5
        & 94.8 \\
        \midrule

        \multicolumn{9}{@{}l}{\textit{BEV-based methods}} \\
        \addlinespace[2pt]

        UniAD~\cite{uniad}
        & CVPR'23
        & 6$\times$ Cam
        & 97.8 & 91.9 & 92.9 & \best{100.0} & 78.8 & 83.4 \\

        TransFuser~\cite{trans}
        & TPAMI'23
        & 3$\times$ Cam + L
        & 97.7 & 92.8 & 92.8 & \best{100.0} & 79.2 & 84.0 \\

        PARA-Drive~\cite{pdrive}
        & CVPR'24
        & 6$\times$ Cam
        & 97.9 & 92.4 & 93.0 & 99.8 & 79.3 & 84.0 \\

        LAW~\cite{law}
        & ICLR'25
        & 1$\times$ Cam
        & 96.4 & 95.4 & 88.7 & 99.9 & 81.7 & 84.6 \\

        Hydra-MDP~\cite{hydra}
        & arXiv'24
        & 3$\times$ Cam + L
        & 98.3 & 96.0 & 94.6 & \best{100.0} & 78.7 & 86.5 \\

        DiffusionDrive~\cite{diffusiondrive}
        & CVPR'25
        & 3$\times$ Cam + L
        & 98.2 & 96.2 & 94.7 & \best{100.0} & 82.2 & 88.1 \\

        WoTE~\cite{wote}
        & ICCV'25
        & 3$\times$ Cam + L
        & 98.5 & 96.8 & 94.4 & 99.9 & 81.9 & 88.3 \\

        \midrule
        \multicolumn{9}{@{}l}{\textit{Normal view methods}} \\
        \addlinespace[2pt]

        AutoVLA~\cite{autovla}
        & NeurIPS'25
        & 3$\times$ Cam
        & 98.4 & 95.6 & 98.0 & 99.9 & 81.9 & 89.1 \\

        ReCogDrive~\cite{recogdrive}
        & arXiv'25
        & 3$\times$ Cam
        & 98.2 & 97.8 & 95.2 & 99.8 & 83.5 & 89.6 \\

        DriveVLA-W0$^{\dagger}$~\cite{drivevla}
        & arXiv'25
        & 1$\times$ Cam
        & 98.7 & \best{99.1} & 95.3 & 99.3 & 83.3 & 90.2 \\

        AutoVLA$^{\ddagger}$~\cite{autovla}
        & NeurIPS'25
        & 3$\times$ Cam
        & 99.1 & 97.1 & \second{97.1}
        & \best{100.0} & 87.6 & 92.1 \\

        DriveVLA-W0$^{\ddagger}$~\cite{drivevla}
        & arXiv'25
        & 1$\times$ Cam
        & \second{99.3} & 97.4 & 97.0
        & 99.9 & \second{88.3} & \second{93.0} \\

        \addlinespace[1pt]
        \rowcolor{oursblue}
        \textbf{Drive-HWM (Ours)}
        & --
        & 1$\times$ Cam
        & \best{99.6}
        & \second{99.0}
        & \best{98.5}
        & \best{100.0}
        & \best{89.0}
        & \best{93.8} \\

        \bottomrule
    \end{tabular}

    \vspace{3pt}

    \begin{minipage}{0.96\textwidth}
        \footnotesize
        \raggedright
        \textbf{Sensors:}
        1$\times$ Cam denotes a single front-view camera;
        3$\times$/6$\times$ Cam denotes surround-view cameras;
        L denotes LiDAR.
        $^{\dagger}$ Query-based action expert with multiple trajectory anchors
        following~\cite{hydra}.
        $^{\ddagger}$ Autoregressive action expert with the best-of-$N$
        strategy ($N=6$) following~\cite{autovla}.
    \end{minipage}
\end{table*}

\begin{table*}[t]
    \centering
    \caption{
        \textbf{Comparison with state-of-the-art methods on NAVSIM v2
        using the extended evaluation metrics.}
        The best and second-best results are highlighted in
        \textbf{bold} and \underline{underlined}, respectively.
        Our method is highlighted in light blue.
    }
    \label{tab:navsim_v2_sota}

    \small
    \setlength{\tabcolsep}{4.8pt}
    \renewcommand{\arraystretch}{1.15}

    \begin{tabular}{@{}lrrrrrrrrrr@{}}
        \toprule
        \textbf{Method}
        & \textbf{NC$\uparrow$}
        & \textbf{DAC$\uparrow$}
        & \textbf{DDC$\uparrow$}
        & \textbf{TLC$\uparrow$}
        & \textbf{EP$\uparrow$}
        & \textbf{TTC$\uparrow$}
        & \textbf{LK$\uparrow$}
        & \textbf{HC$\uparrow$}
        & \textbf{EC$\uparrow$}
        & \textbf{EPDMS$\uparrow$} \\
        \midrule

        Ego Status
        & 93.1
        & 77.9
        & 92.7
        & 99.6
        & 86.0
        & 91.5
        & 89.4
        & \best{98.3}
        & 85.4
        & 64.0 \\

        TransFuser~\cite{trans}
        & 96.9
        & 89.9
        & 97.8
        & 99.7
        & 87.1
        & 95.4
        & 92.7
        & \best{98.3}
        & 87.2
        & 76.7 \\

        HydraMDP++~\cite{hydra}
        & 97.2
        & 97.5
        & \best{99.4}
        & 99.6
        & 83.1
        & 96.5
        & 94.4
        & \second{98.2}
        & 70.9
        & 81.4 \\

        DriveSupervisor~\cite{drivesuprim}
        & 97.5
        & 96.5
        & \best{99.4}
        & 99.6
        & \second{88.4}
        & 96.6
        & 95.5
        & \best{98.3}
        & 77.0
        & 83.1 \\

        ARTEMIS~\cite{artemis}
        & 98.3
        & 95.1
        & \second{98.6}
        & \second{99.8}
        & 81.5
        & 97.4
        & \second{96.5}
        & \best{98.3}
        & --
        & 83.1 \\

        DiffusionDrive~\cite{diffusiondrive}
        & 98.2
        & 95.9
        & \best{99.4}
        & \second{99.8}
        & 87.5
        & 97.3
        & \best{96.8}
        & \best{98.3}
        & \best{87.7}
        & 84.5 \\

        DriveVLA-W0~\cite{drivevla}
        & \second{98.5}
        & \second{99.1}
        & 98.0
        & 99.7
        & 86.4
        & \second{98.1}
        & 93.2
        & 97.9
        & 58.9
        & \second{86.1} \\

        \midrule

        \rowcolor{oursblue}
        \textbf{Drive-HWM (Ours)}
        & \best{98.9}
        & \best{99.5}
        & \best{99.4}
        & \best{99.9}
        & \best{88.5}
        & \best{98.7}
        & \second{96.5}
        & \best{98.3}
        & \second{87.5}
        & \best{86.4} \\

        \bottomrule
    \end{tabular}

    \vspace{3pt}

    \begin{minipage}{0.98\textwidth}
        \footnotesize
        \raggedright
        \textbf{Metrics:}
        NC: No At-fault Collision;
        DAC: Drivable Area Compliance;
        DDC: Driving Direction Compliance;
        TLC: Traffic Light Compliance;
        EP: Ego Progress;
        TTC: Time to Collision;
        LK: Lane Keeping;
        HC: History Comfort;
        EC: Extended Comfort;
        EPDMS: Extended Predictive Driver Model Score.
        ``--'' denotes an unavailable result.
    \end{minipage}
\end{table*}

\section{Experiment}

This section presents the experimental evaluation of Drive-HWM. We first introduce the datasets, evaluation metrics, and implementation details. We then quantitatively compare Drive-HWM with state-of-the-art methods to assess its overall driving performance. Next, we provide qualitative visualizations to analyze the long-horizon anticipation and action-generation capabilities of the proposed slow--fast hierarchy. Finally, we conduct comprehensive ablation studies to examine the contribution of each component and validate the effectiveness of our design choices.

\subsection{Implementation Details}

\textbf{Datasets and evaluation.}
We pretrain our model on nuPlan~\cite{nuplan} and fine-tune it on
NAVSIM~\cite{navsim}, which is built upon the OpenScene
dataset~\cite{openscene}. We report results under both NAVSIM v1 and
NAVSIM v2~\cite{nav2}. NAVSIM v1 evaluates No At-fault Collision (NC),
Drivable Area Compliance (DAC), Time-to-Collision (TTC), Comfort (C.), and Ego
Progress (EP), and summarizes performance using
$\mathrm{PDMS}=\mathrm{NC}\times\mathrm{DAC}\times
(5\mathrm{EP}+5\mathrm{TTC}+2\mathrm{C})/12$.
NAVSIM v2 additionally considers Driving Direction Compliance (DDC), Traffic
Light Compliance (TLC), Lane Keeping (LK), History Comfort (HC), and Extended
Comfort (EC), with
$\mathrm{EPDMS}=\mathrm{NC}\times\mathrm{DAC}\times\mathrm{DDC}\times
\mathrm{TLC}\times
(5\mathrm{EP}+5\mathrm{TTC}+2\mathrm{LK}+2\mathrm{HC}+2\mathrm{EC})/16$.

\textbf{Training details.}
We set the slow-model update interval and prediction horizon to $N=K=8$.
Accordingly, the slow world model is updated every eight timesteps and predicts
Dynamic-Aware Latents for the next eight timesteps, while the fast world model
runs at every timestep. We adopt a two-stage training strategy: the backbone,
slow world model, and fast world model are first jointly pretrained on nuPlan
for 10K steps using the full objective $\mathcal{L}_{\mathrm{total}}$; the complete model is then fine-tuned on NAVSIM for 6K
steps using the action prediction loss. We resize input images to
$256\times144$ and train the model on eight NVIDIA A100 GPUs with a global
batch size of 48. We use AdamW with an initial learning rate of
$2\times10^{-4}$ and a cosine learning-rate schedule.

\begin{figure*}[t]
\centering
\includegraphics[height=5.5cm,width=1.0\linewidth]{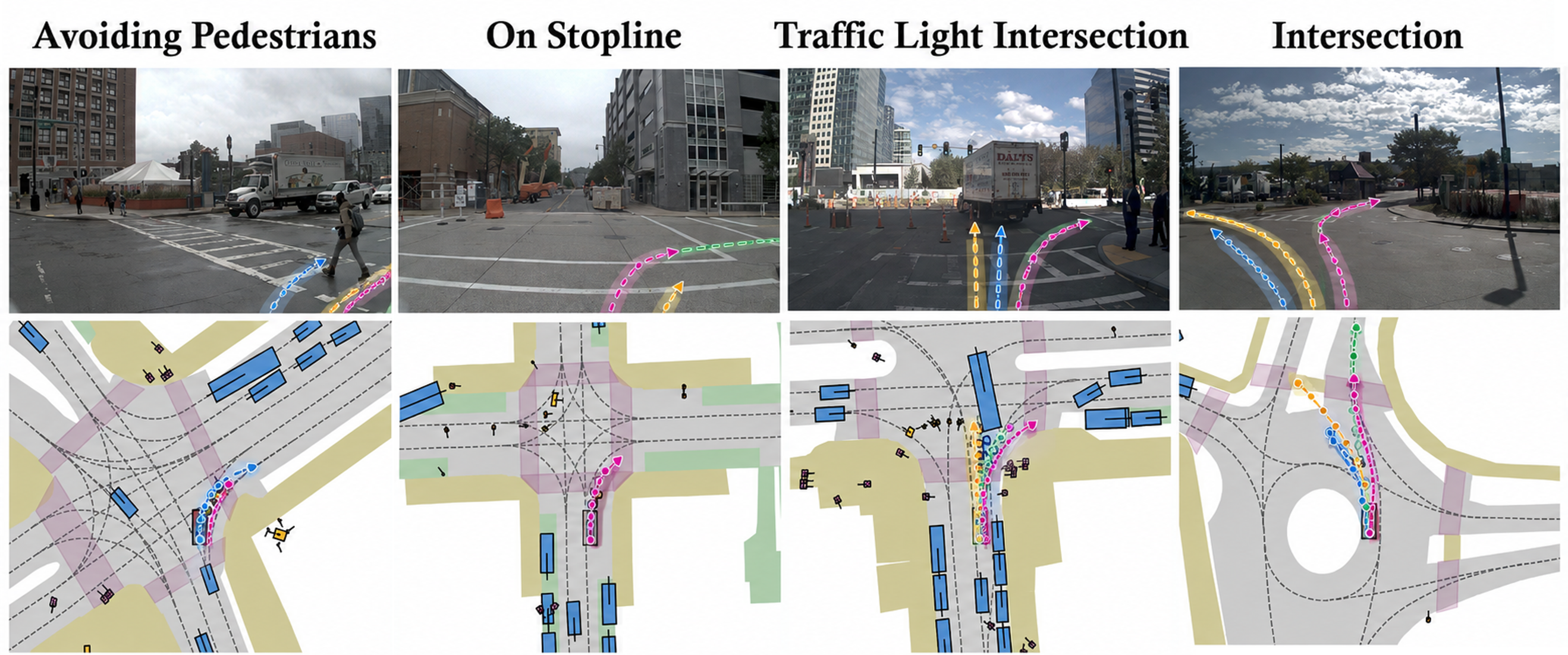}
\caption{\textbf{Qualitative comparison of trajectories from different models in the
front-facing camera and bird's-eye view across diverse driving scenarios.}
Trajectories are shown for:
{\color{green!70!black}\rule{1.2em}{2pt}}~Human Trajectory,
{\color{magenta}\rule{1.2em}{2pt}}~Drive-HWM,
{\color{cyan}\rule{1.2em}{2pt}}~DriveVLA-W0,
{\color{orange}\rule{1.2em}{2pt}}~Transfuser.}
\label{fig:qualitative}
% \label{fig:vis_traj}
\vspace{-0.2in}
\end{figure*}

\subsection{Quantitative Comparison}

Tables~\ref{tab:navsim_v1_sota} and~\ref{tab:navsim_v2_sota} compare Drive-HWM with state-of-the-art methods under the two NAVSIM protocols. On NAVSIM v1, Drive-HWM achieves the highest overall PDMS of 93.3 using only a single front-view camera and obtains the best or tied-best results on NC, TTC, comfort, and EP. In particular, compared with DriveVLA-W0~\cite{drivevla}, which adopts the same VLA backbone but uses a single world model, Drive-HWM improves NC, DAC, TTC, comfort, EP, and PDMS by 0.2, 1.0, 1.0, 0.1, 0.2, and 0.3 points, respectively, over its strongest overall variant. The improvements in NC and TTC, which primarily measure collision avoidance and temporal safety margins, demonstrate that the slow branch can anticipate long-term scene evolution and potential risks, while the fast branch enables timely reactions to rapidly changing local conditions. Similarly, the gain in EP indicates that such long-horizon reasoning allows the agent to make safer yet less overly conservative progress. Although our DAC of 98.4 is slightly lower than the best individual result of 99.1, it remains highly comparable; unlike the aggregate metrics, DAC mainly evaluates instantaneous adherence to drivable-area boundaries and is therefore more sensitive to local geometric precision than to long-term planning quality. Overall, the leading PDMS confirms that the proposed slow--fast hierarchy achieves a more favorable balance among safety, compliance, progress, and comfort than methods relying on a single-timescale world model.

On the more comprehensive NAVSIM v2 benchmark, Drive-HWM again ranks first in the overall EPDMS, achieving 86.2, and obtains the best or tied-best performance on seven of the ten metrics, including NC, DAC, DDC, TLC, TTC, HC, and EPDMS. Compared with DriveVLA-W0~\cite{drivevla}, Drive-HWM improves all ten reported metrics, with particularly substantial gains of 1.4 points in DDC, 1.7 points in EP, 1.5 points in LK, and 27.5 points in EC. The leading results on NC and TTC further verify the benefit of slow, long-horizon prediction for anticipating future hazards, whereas the improvements in DAC, DDC, and TLC show that the fast branch can convert high-level plans into responsive and rule-compliant actions. Drive-HWM also achieves a comparable EP of 88.1, only 0.3 points below the best result, indicating that its stronger safety performance is not obtained through excessively conservative driving. Its LK and EC results are competitive but not the best: LK emphasizes fine-grained lane-centering accuracy, while EC is particularly sensitive to short-term acceleration and steering variations. The fast branch may introduce occasional local corrections when responding to dynamic risks, slightly affecting these two metrics, whereas the slow branch prioritizes globally safe and temporally consistent execution over optimizing isolated measures of geometric tracking or instantaneous smoothness. 

Taken together, the best aggregate scores under both protocols, broad improvements over the same-backbone baseline, and strong performance with only a single front-view camera demonstrate that Drive-HWM provides the strongest overall trade-off, validating the effectiveness of hierarchical slow--fast world modeling for reliable long-horizon planning and responsive short-horizon control.

\begin{table}[t]
    \centering
    \caption{\textbf{Computational efficiency comparison.}
    Latency is measured on a single NVIDIA H200 GPU with batch size 1.
    $T_s$ and $T_f$ denote the inference latency of the slow and fast
    models, respectively. Since the slow model is executed every $N$
    steps, the amortized latency is computed as
    $T_{\mathrm{avg}} = T_f + T_s/N$.
    $T_{\mathrm{peak}}$ denotes the latency at a slow-model update step.}
    \label{tab:computational_efficiency}
    
    \small
    \setlength{\tabcolsep}{3.8pt}
    \renewcommand{\arraystretch}{1.12}
    
    \begin{tabular}{@{}lcccccc@{}}
        \toprule
        \textbf{Method}
        & \textbf{$N$}
        & \textbf{$T_s$}
        & \textbf{$T_f$}
        & \textbf{$T_{\mathrm{peak}}$}
        & \textbf{$T_{\mathrm{avg}}$}
        & \textbf{PDMS$\uparrow$} \\
        & & \multicolumn{4}{c}{\textbf{Latency (ms)$\downarrow$}} & \\
        \midrule
        
        DriveVLA-W0~\cite{drivevla}
        & -- 
        & -- 
        & 117.8 
        & 117.8
        & 117.8 
        & 93.0 \\
        
        Fast Model Only
        & -- 
        & -- 
        & \textbf{81.6}
        & \textbf{81.6}
        & \textbf{81.6}
        & 93.0 \\
        
        \rowcolor{oursblue}
        \textbf{Drive-HWM (Ours)}
        & 8
        & 25.6
        & 81.6
        & 107.2
        & \underline{84.8}
        & \textbf{93.8} \\
        
        \bottomrule
    \end{tabular}
\end{table}

\textbf{Computational Efficiency.}
We evaluate the inference efficiency of Drive-HWM on a single NVIDIA H200 GPU with batch size 1. As shown in Table~\ref{tab:computational_efficiency}, the fast world model requires 81.6\,ms per control step, while the slow world model takes 25.6\,ms and is executed only once every $N=8$ steps. This amortizes the slow-branch overhead to 3.2\,ms per step, resulting in an average latency of 84.8\,ms. Compared with DriveVLA-W0~\cite{drivevla} at 117.8\,ms, Drive-HWM reduces latency by approximately 28\% while improving PDMS from 93.0 to 93.3. These results show that the hierarchical slow--fast design introduces long-horizon predictive reasoning with only marginal computational overhead.
% \noindent\textbf{Computational Efficiency.}
% We further evaluate the inference efficiency of Drive-HWM on a single
% NVIDIA H200 GPU with batch size 1.
% As shown in Table~\ref{tab:computational_efficiency}, the fast world
% model requires 81.6\,ms for each control step, while one execution of
% the slow world model takes 25.6\,ms.
% Importantly, the slow model is evaluated only once every $N=8$ steps,
% and its predicted Dynamic-Aware Latents are reused by the subsequent
% fast-model steps.
% Therefore, the computational cost of the slow branch is amortized to
% only 3.2\,ms per control step, resulting in an average inference
% latency of 84.8\,ms for the complete Drive-HWM.

% Compared with DriveVLA-W0, which requires 117.8\,ms per inference,
% Drive-HWM reduces the average latency by approximately 28\% while
% improving PDMS from 93.0 to 93.3.
% Moreover, compared with the fast-only variant, introducing the
% long-horizon slow world model incurs only 3.2\,ms additional
% amortized latency, yet consistently improves driving performance.
% Even at slow-model update steps, the peak latency is 107.2\,ms,
% remaining below the latency of the DriveVLA-W0 baseline.
% These results demonstrate that the proposed hierarchical design
% introduces long-horizon predictive reasoning with only marginal
% per-step computational overhead, providing a favorable trade-off
% between driving performance and inference efficiency.

\subsection{Qualitative Comparison}
Fig.\ref{fig:qualitative} provides qualitative comparisons in four representative and challenging scenarios, with trajectories visualized in both the front-camera view and the bird's-eye view. 

In the pedestrian-avoidance scenario, Drive-HWM predicts a curved trajectory that closely follows the human trajectory while maintaining sufficient clearance from the crossing pedestrian. In contrast, the baseline trajectories exhibit larger spatial deviations, suggesting less accurate modeling of the interaction between ego motion and the pedestrian’s future occupancy. At the intersection, Drive-HWM accurately captures the intended turning direction and produces a trajectory nearly overlapping with the human demonstration. TransFuser\cite{trans} and DriveVLA-W0~\cite{drivevla}, however, deviate noticeably toward alternative paths, indicating ambiguity in their understanding of the long-term route intention and intersection topology. In the stop-line scenario, our method correctly anticipates the maneuver beyond the stop line and generates a geometrically coherent trajectory consistent with the reference behavior, whereas the baselines show less accurate directional alignment.

Finally, at the traffic-light-controlled intersection, Drive-HWM follows the human trajectory through the intended turning corridor, while the competing methods tend to continue toward the center of the intersection and fail to fully capture the upcoming route transition. This distinction is particularly important because the current observation alone provides limited evidence about the complete maneuver, requiring the planner to reason jointly about traffic signals, road topology, and future route evolution. Across all four scenarios, Drive-HWM exhibits stronger agreement with human driving behavior in terms of maneuver intention, lane-level geometry, obstacle avoidance, and traffic-rule compliance. 

These results qualitatively demonstrate the benefit of the proposed hierarchical slow--fast world modeling: the slow world model anticipates long-horizon scene evolution and preserves high-level route and interaction intentions, while the fast world model grounds this predicted future context in the latest observation to make timely local corrections. Their complementary operation enables Drive-HWM to avoid the short-sighted or directionally inconsistent trajectories produced by single-timescale baselines, resulting in safer, more accurate, and more human-like trajectory generation across diverse driving conditions.

\begin{table}[t]
    \centering
    \caption{
        \textbf{Ablation studies on the hierarchical slow--fast architecture
        and future prediction horizon.}
        The best and second-best results are highlighted in bold and
        underlined, respectively.
    }
    \label{tab:hierarchy-horizon-ablation}

    \small
    \setlength{\tabcolsep}{4.2pt}
    \renewcommand{\arraystretch}{1.15}

    \begin{tabular}{@{}lcccrrr@{}}
        \toprule
        \textbf{Configuration}
        & \textbf{Slow}
        & \textbf{Fast}
        & \textbf{$K$}
        & \textbf{NC$\uparrow$}
        & \textbf{DAC$\uparrow$}
        & \textbf{PDMS$\uparrow$} \\
        \midrule

        Fast only
        & --
        & $\checkmark$
        & --
        & \second{99.3}
        & 97.4
        & 93.0 \\

        Slow only
        & $\checkmark$
        & --
        & 8
        & 98.2
        & 97.1
        & 90.2 \\

        \midrule

        Drive-HWM
        & $\checkmark$
        & $\checkmark$
        & 4
        & 99.0
        & 98.0
        & 93.0 \\

        \rowcolor{oursblue}
        \textbf{Drive-HWM (Ours)}
        & $\checkmark$
        & $\checkmark$
        & \textbf{8}
        & \best{99.6}
        & \best{99.0}
        & \best{93.8} \\

        Drive-HWM
        & $\checkmark$
        & $\checkmark$
        & 12
        & 99.2
        & \second{98.4}
        & \second{93.2} \\

        \bottomrule
    \end{tabular}

    \vspace{3pt}

    \begin{minipage}{0.98\linewidth}
        \footnotesize
        \raggedright
        ``Slow'' and ``Fast'' denote the slow world model and fast
        action model, respectively. $K$ denotes the planning horizon
        of the slow world model. The configuration with $K=8$ is used
        as our default setting.
    \end{minipage}
\end{table}

\begin{table}[t]
    \centering
    \caption{
        \textbf{Ablation study on the backbone architectures of the slow
        and fast models.}
        The best and second-best results within each group are highlighted
        in bold and underlined, respectively.
    }
    \label{tab:backbone-ablation}

    \small
    \setlength{\tabcolsep}{7.5pt}
    \renewcommand{\arraystretch}{1.15}

    \begin{tabular}{@{}lccc@{}}
        \toprule
        \textbf{Backbone}
        & \textbf{NC$\uparrow$}
        & \textbf{DAC$\uparrow$}
        & \textbf{PDMS$\uparrow$} \\
        \midrule

        \rowcolor{sectiongray}
        \multicolumn{4}{@{}l}{
            \textit{\textbf{Slow World Model Backbone}}
            \quad (Fast model: Emu3)
        } \\

        CogVideo~\cite{cogvideo}
        & 98.8
        & 98.1
        & 93.0 \\

        WAN~\cite{wan}
        & \underline{99.0}
        & \underline{98.4}
        & \underline{93.2} \\

        \rowcolor{oursblue}
        \textbf{V-JEPA~\cite{vjepa} (Ours)}
        & \textbf{99.6}
        & \textbf{99.0}
        & \textbf{93.8} \\

        \midrule

        \rowcolor{sectiongray}
        \multicolumn{4}{@{}l}{
            \textit{\textbf{Fast Action Model Backbone}}
            \quad (Slow model: V-JEPA)
        } \\

        LLaVA-OneVision~\cite{li2024llava}
        & 99.0
        & 98.4
        & \underline{93.5} \\

        Qwen2.5-VL~\cite{bai2025qwen3}
        & \underline{99.1}
        & \underline{98.6}
        & 93.3 \\

        \rowcolor{oursblue}
        \textbf{Emu3~\cite{emu3} (Ours)}
        & \textbf{99.6}
        & \textbf{99.0}
        & \textbf{93.8} \\

        \bottomrule
    \end{tabular}

    \vspace{3pt}
    \begin{minipage}{0.96\linewidth}
        \footnotesize
        \raggedright
        When ablating the slow world model backbone, the fast action model
        is fixed as Emu3. When ablating the fast action model backbone,
        the slow world model is fixed as V-JEPA.
    \end{minipage}
    \vspace{-0.2in}
\end{table}

\subsection{Ablation Study}

\paragraph{\textbf{Impact of the  Hierarchical Slow--Fast World Modeling Architecture}}  
The slow--fast hierarchy is a key design of Drive-HWM, enabling the model to combine long-horizon scene anticipation with observation-grounded action generation. In this part, we investigate its effectiveness by removing either the slow or fast component and further study the influence of the prediction horizon $K$. As shown in Table~\ref{tab:hierarchy-horizon-ablation}, retaining only the fast model decreases NC, DAC, and PDMS from 99.5, 98.4, and 93.3 to 99.3, 97.4, and 93.0, respectively. Without the slow model, the policy mainly reacts to the current observation and lacks sufficient foresight for anticipating future road structures and potential risks. The slow-only variant leads to a larger PDMS drop of 3.1 points because long-horizon predictions alone cannot promptly adapt to newly observed changes or correct prediction errors during execution. Combining the two components achieves the best overall performance: the slow model provides temporally coherent planning guidance, while the fast model continuously grounds this guidance in the latest observation for responsive control. We also compare prediction horizons of $K=4$, $8$, and $12$. A short horizon provides insufficient future context, whereas an excessively long horizon introduces greater prediction uncertainty. Consequently, $K=8$ achieves the best balance between long-term foresight and prediction reliability, yielding the highest NC, DAC, and PDMS scores. These results verify the effectiveness of the proposed slow--fast decomposition and justify our choice of an eight-frame prediction horizon.

\paragraph{\textbf{Impact of the  Foundation World Model Backbone Choice}}  
The choice of foundation model backbones is important for Drive-HWM, as it directly affects both future-scene prediction in the slow world model and observation-grounded action generation in the fast model. To investigate its impact, we separately replace the backbone of each component while keeping the other component fixed. Specifically, we compare V-JEPA with the generative video models CogVideo and WAN for the slow world model, and compare Emu3 with alternative vision-language backbones for the fast model. As shown in Table~\ref{tab:backbone-ablation}, V-JEPA achieves the best overall performance among the slow-model backbones. Compared with generative video models, its latent-space predictive objective focuses more directly on temporally meaningful scene dynamics while avoiding the unnecessary complexity of reconstructing low-level visual details, making it more suitable for long-horizon driving anticipation. For the fast model, Emu3 achieves the best performance across the representative metrics, demonstrating its stronger ability to integrate the latest visual observation with the future context predicted by the slow model. Overall, the results validate V-JEPA and Emu3 as effective and complementary backbone choices for long-horizon world modeling and responsive action generation, respectively.

\begin{figure}[h]
\centering
\includegraphics[width=1.0\linewidth]{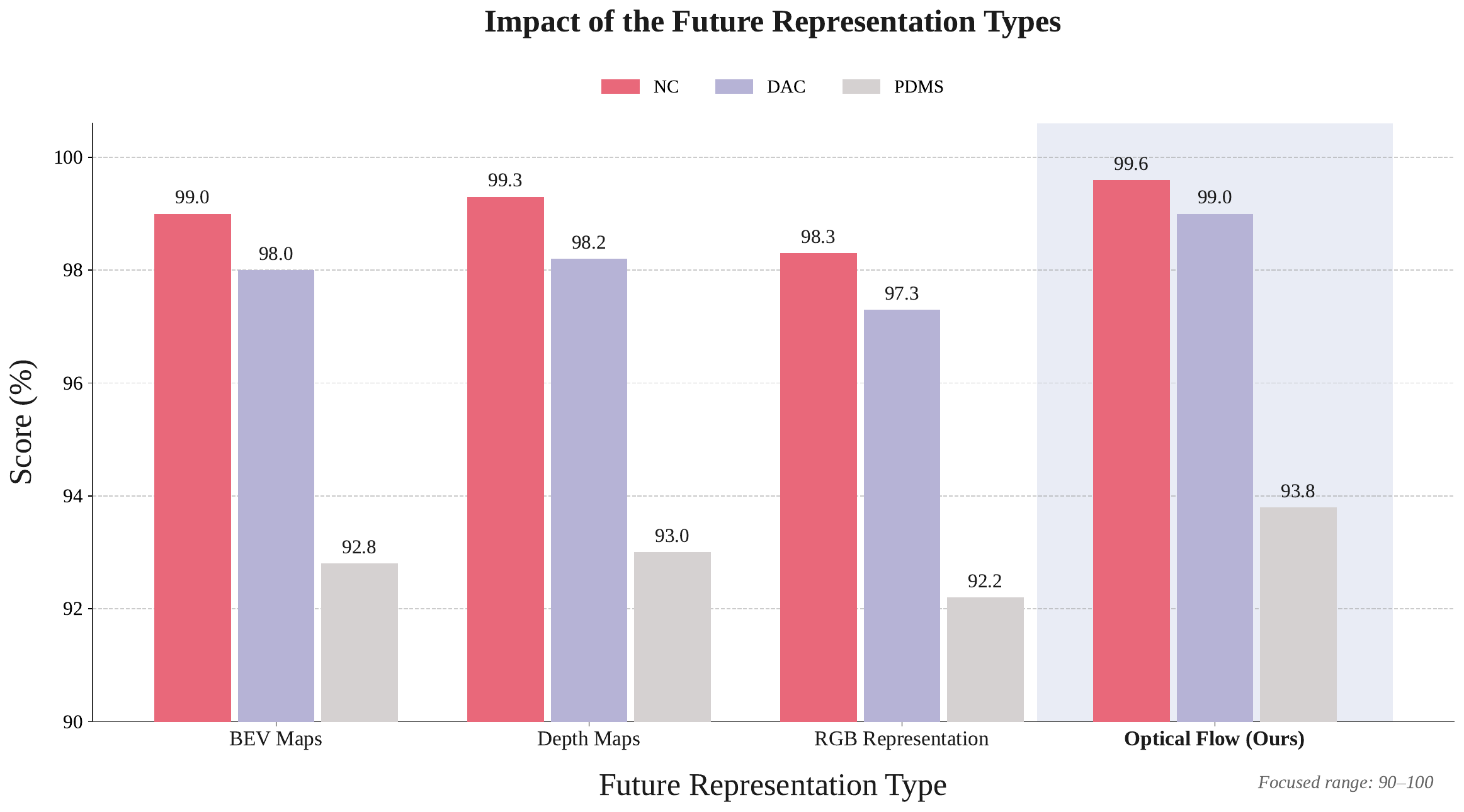}
\caption{Quantitative comparison of different prediction targets for supervising the slow world model. Optical-flow supervision consistently achieves the highest NC, DAC, and PDMS scores, demonstrating that explicitly modeling dense scene dynamics provides more informative future representations for downstream driving decisions and action generation.}
\label{fig: representation_types}
\end{figure}

\begin{figure}[h]
\centering
\includegraphics[width=1.0\linewidth]{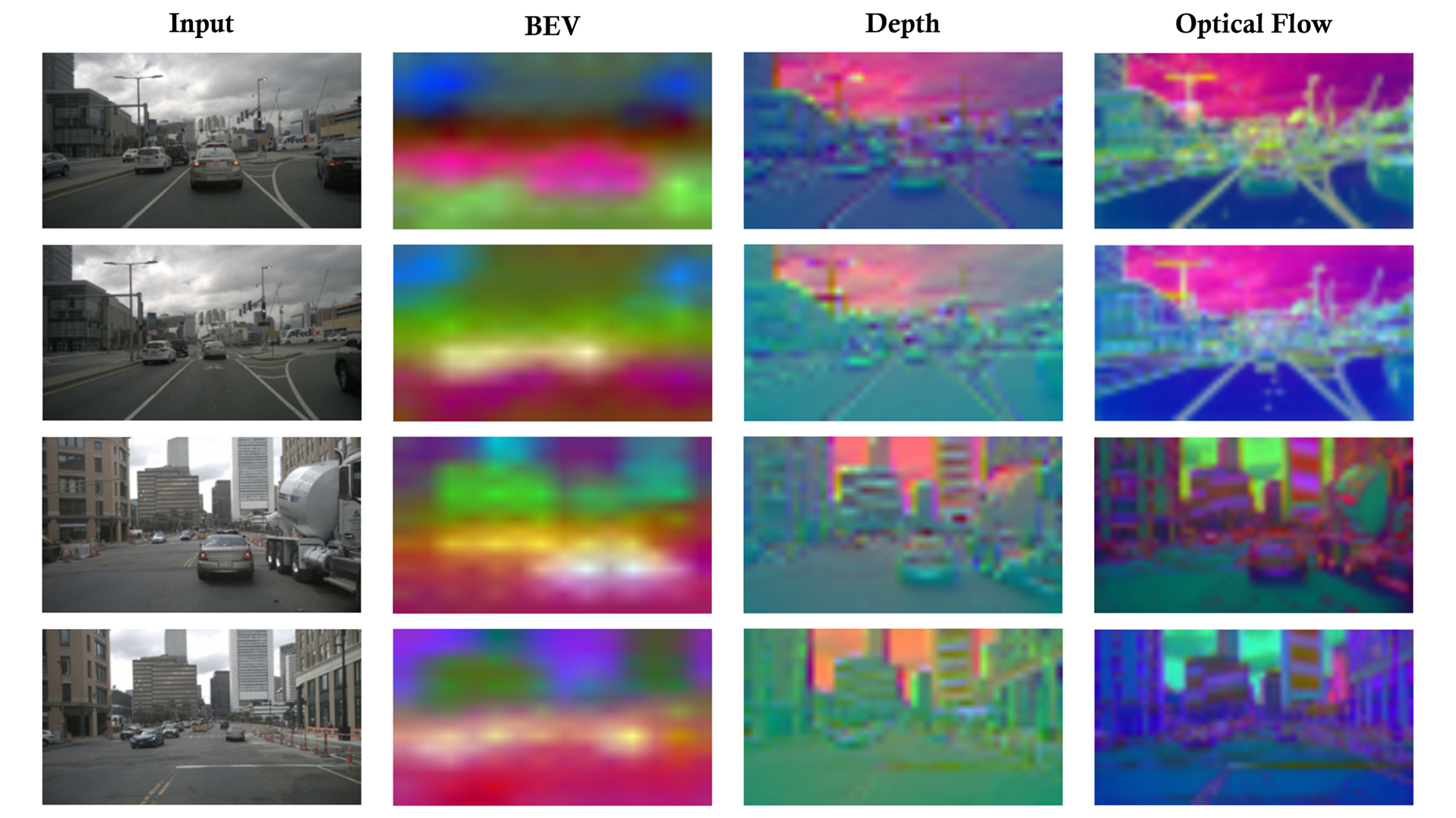}
\caption{Visualization of different future representations predicted by the slow world model. Compared with BEV and depth representations, optical flow provides denser and more motion-discriminative cues, explicitly capturing scene dynamics and object displacements that are essential for modeling future evolution in autonomous driving.}
\label{fig:supervision_visualization}
\end{figure}

\paragraph{ \textbf{Impact of the  Prediction Target of the Slow Model}} The choice of prediction target is crucial for the slow world model, as it determines whether the learned latent representation can effectively capture the scene dynamics relevant to downstream action generation. To investigate this design, we supervise the slow model using four future representations---BEV maps, depth maps, RGB observations, and optical flow---while keeping the remaining model configurations unchanged. As shown in Fig.~\ref{fig: representation_types}, RGB prediction yields the lowest performance because reconstructing appearance details may distract the model from learning planning-relevant motion patterns. BEV and depth supervision provide stronger geometric and spatial cues, but only implicitly characterize temporal changes and object movements. In contrast, optical flow achieves the best results. As illustrated in Fig.~\ref{fig:supervision_visualization}, optical flow offers dense and motion-discriminative supervision that explicitly describes pixel-level displacements and dynamic scene evolution. Consequently, its learned latent representation preserves richer motion information, providing more informative future context for driving decisions and action generation. These results demonstrate that optical flow is a more effective prediction target for dynamics-aware world modeling in autonomous driving.

\paragraph{\textbf{Motion-Aware Representation Probing}}  Learning motion-aware representations is essential for the slow world model, since its latent features should encode future scene dynamics rather than merely preserve static appearance or geometry. So we may wonder: does the slow world model learn dynamic-aware latent? To examine whether such information is captured, we freeze the slow world models trained with different prediction targets and train lightweight linear probes on their latent representations to predict future ego motion and temporal motion consistency. The slow world models remain fixed throughout probing, ensuring that the results reflect motion information already encoded in their representations. As shown in Table~\ref{tab:motion-probe}, the optical-flow latent achieves the highest probing accuracy on both tasks, outperforming the second-best RGB latent by 6.9 points on future ego-motion prediction and 6.7 points on motion consistency. Unlike BEV, depth, and RGB targets, optical flow directly provides dense displacement cues associated with both ego motion and dynamic scene evolution. Consequently, optical-flow supervision encourages the slow world model to learn a more general dynamic-aware latent representation, which provides informative dynamic priors for downstream driving decisions and action generation.

\begin{table}[t]
    \centering
    \caption{\textbf{Linear probing of motion information encoded in the slow-world-model representations.}
    FEM and MC denote Future Ego Motion and Motion Consistency, respectively.
    The best and second-best results are highlighted in bold and underlined.}
    \label{tab:motion-probe}
    
    \small
    \setlength{\tabcolsep}{7pt}
    \renewcommand{\arraystretch}{1.12}
    
    \begin{tabular}{@{}lcc@{}}
        \toprule
        \textbf{Probed Latent}
        & \textbf{FEM$\uparrow$}
        & \textbf{MC$\uparrow$} \\
        \midrule
        
        BEV
        & 71.2 & 63.8 \\
        
        Depth
        & 74.5 & 66.9 \\
        
        RGB
        & \underline{76.8} & \underline{69.4} \\
        
        \rowcolor{oursblue}
        \textbf{Optical Flow (Ours)}
        & \textbf{83.7} & \textbf{76.1} \\
        
        \bottomrule
    \end{tabular}
\end{table}

\begin{table}[t]
    \centering
    \caption{
        \textbf{Effect of slow-to-fast conditioning strategies.} GCA denotes gated cross-attention. The latent representation produced by the slow world model is injected into the fast action
        model using different conditioning mechanisms.
    }
    \label{tab:plan-fusion}

    \small
    \setlength{\tabcolsep}{6pt}
    \renewcommand{\arraystretch}{1.15}

    \begin{tabular}{@{}lccc@{}}
        \toprule
        \textbf{Conditioning}
        & \textbf{NC$\uparrow$}
        & \textbf{DAC$\uparrow$}
        & \textbf{PDMS$\uparrow$} \\
        \midrule

        Concatenation
        & 98.2 & 97.7 & 92.5 \\

        Cross-Attention
        & 98.7 & 98.0 & 93.0 \\

        GCA~\cite{alayrac2022flamingo}
        & 98.9 & 98.5 & 93.1 \\

        AdaLN~\cite{guo2022adaln}
        & \underline{99.2} & \underline{98.6} & \underline{93.3} \\

        \rowcolor{oursblue}
        \textbf{FiLM~\cite{film} (Ours)}
        & \textbf{99.6}
        & \textbf{99.0}
        & \textbf{93.8} \\

        \bottomrule
    \end{tabular}
\end{table}

\paragraph{\textbf{Impact of Slow-to-Fast Conditioning Strategies}}  Effectively injecting the dynamic-aware latent representation into the fast action model is critical for translating predicted scene dynamics into responsive driving actions. To investigate this design, we compare several slow-to-fast conditioning strategies, including feature concatenation, cross-attention, gated cross-attention, AdaLN, and FiLM, while keeping all other model configurations unchanged. As shown in Table~\ref{tab:plan-fusion}, FiLM achieves the best overall performance, reaching 99.5 NC, 98.4 DAC, and 93.3 PDMS. Direct concatenation provides only limited interaction between the slow and fast representations, while attention-based methods introduce additional computational complexity and may not consistently preserve the global planning context. In contrast, FiLM transforms the dynamic-aware latent representation into feature-wise affine parameters that adaptively modulate the intermediate features of the fast model. This enables future prediction guidance to be injected throughout action generation without disrupting the observation-grounded representations.

\paragraph{\textbf{Impact of the Next-frame Visual Supervision in the Fast Model}}   The auxiliary supervision of the fast model is important because it determines what information is preserved for immediate action generation. To study its impact, we compare action-only training with several next-frame prediction targets, including RGB, depth, and optical flow, while keeping all other settings unchanged. As shown in Table~\ref{tab:fast-visual-supervision}, next-frame RGB supervision achieves the best overall performance. Unlike the slow world model, whose primary role is to anticipate long-horizon scene dynamics, the fast model focuses on making observation-grounded decisions at the current moment. Its representation must therefore retain comprehensive local evidence, including object semantics, lane markings, traffic signals, spatial layouts, and the states of nearby agents. RGB prediction provides richer and more complete scene information than depth or optical flow, which primarily emphasize geometry or motion, respectively. Moreover, predicting the next RGB frame encourages the fast model to capture immediate visual changes without discarding decision-critical appearance and semantic cues. These results suggest that optical flow is more suitable for learning dynamics-oriented representations in the slow world model, whereas next-frame RGB supervision better supports the fast model in generating accurate and responsive driving actions.

% Required packages:
% \usepackage{booktabs}
% \usepackage[table]{xcolor}

\begin{table}[t]
    \centering
    \caption{
        \textbf{Impact of next-frame visual supervision in the fast model.}
        All variants are trained with action supervision, while differing
        only in the auxiliary target used for next-frame prediction.
        ``None'' denotes training with action supervision only.
    }
    \label{tab:fast-visual-supervision}

    \small
    \setlength{\tabcolsep}{7pt}
    \renewcommand{\arraystretch}{1.15}

    \begin{tabular}{@{}lccc@{}}
        \toprule
        \textbf{Auxiliary Target}
        & \textbf{NC$\uparrow$}
        & \textbf{DAC$\uparrow$}
        & \textbf{PDMS$\uparrow$} \\
        \midrule

        None (Action Only)
        & 99.2 & 98.5 & 93.1 \\

        Next-Optical Flow
        & \underline{99.5} & 98.7 & 93.5 \\

        Next-frame Depth
        & 99.4 & \underline{98.8} & \underline{93.6} \\

        \rowcolor{oursblue}
        \textbf{Next-frame RGB (Ours)}
        & \textbf{99.6} & \textbf{99.0} & \textbf{93.8} \\

        \bottomrule
    \end{tabular}
\end{table}

\begin{figure}[t]
    \centering
    \includegraphics[width=0.5\textwidth]{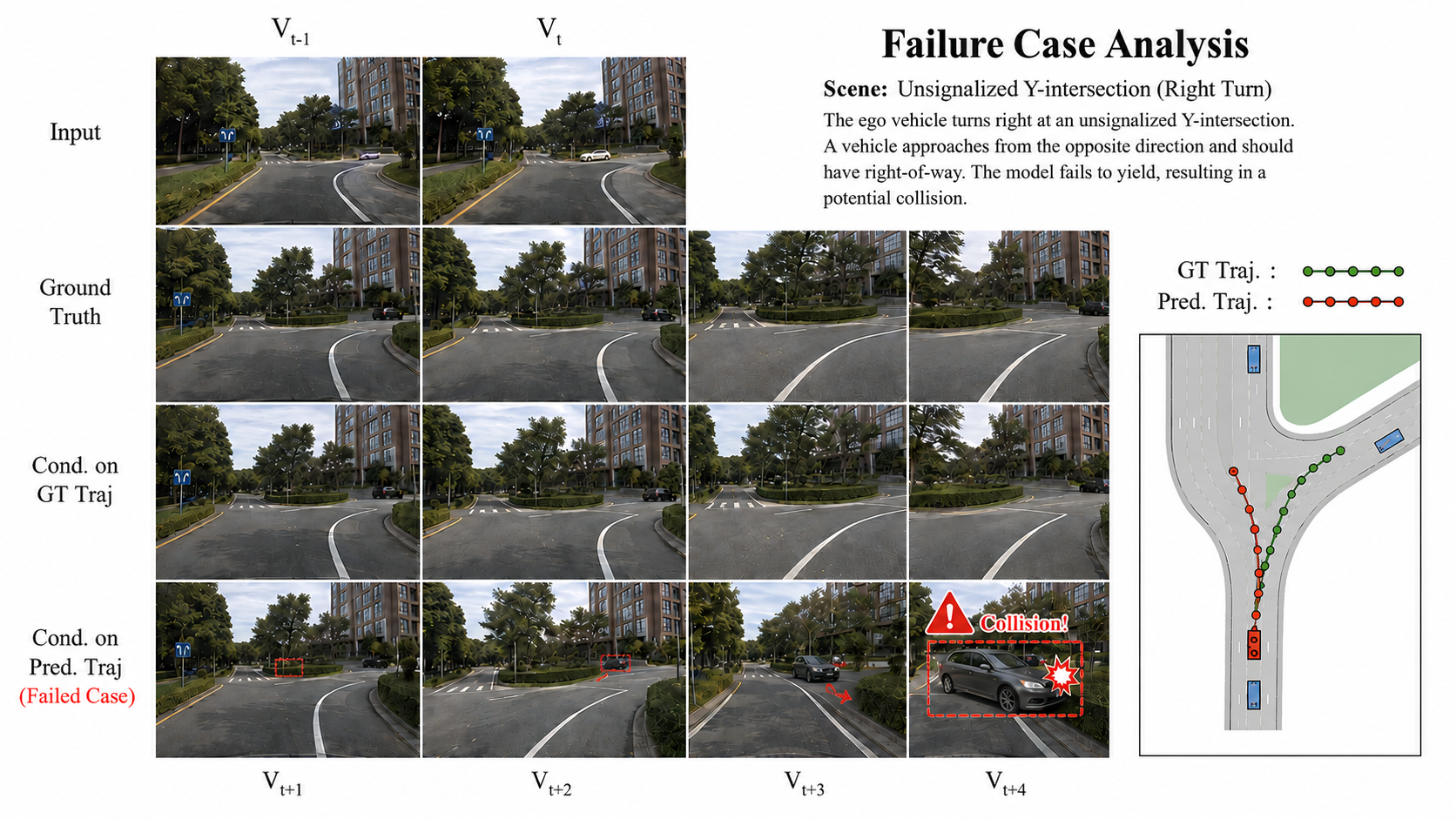}
    \caption{
    \textbf{Failure case analysis of Drive-HWM at an unsignalized Y-intersection.}
    The first row shows the observed input frames, followed by the ground-truth future observations.
    The third row shows future observations conditioned on the ground-truth trajectory, while the last row shows those conditioned on the predicted trajectory.
    In the failed rollout, an approaching vehicle progressively enters the ego vehicle's path, while the predicted trajectory does not yield sufficiently, eventually resulting in a potential collision.
    The bird's-eye-view visualization on the right compares the ground-truth and predicted ego trajectories.
    }
    \label{fig:failure_case}
\end{figure}

\paragraph{\textbf{Failure Analysis}}:
We further analyze a representative failure case at an unsignalized Y-intersection, as illustrated in Fig.~\ref{fig:failure_case}.
In this scenario, the ego vehicle is expected to yield to an approaching vehicle before completing the right turn.
When conditioned on the ground-truth trajectory, the generated future observations remain consistent with the safe maneuver.
In contrast, under the predicted trajectory, the approaching vehicle gradually enters the ego path from $V_{t+1}$ to $V_{t+3}$, while the ego vehicle continues the turn without sufficient yielding, eventually leading to a potential collision at $V_{t+4}$.
This case suggests that Drive-HWM may still struggle in highly interactive scenarios where accurate anticipation of other agents' future motion is critical, motivating better modeling of multimodal futures and interaction uncertainty.

\begin{figure}[t]
    \centering
    \includegraphics[width=\columnwidth]{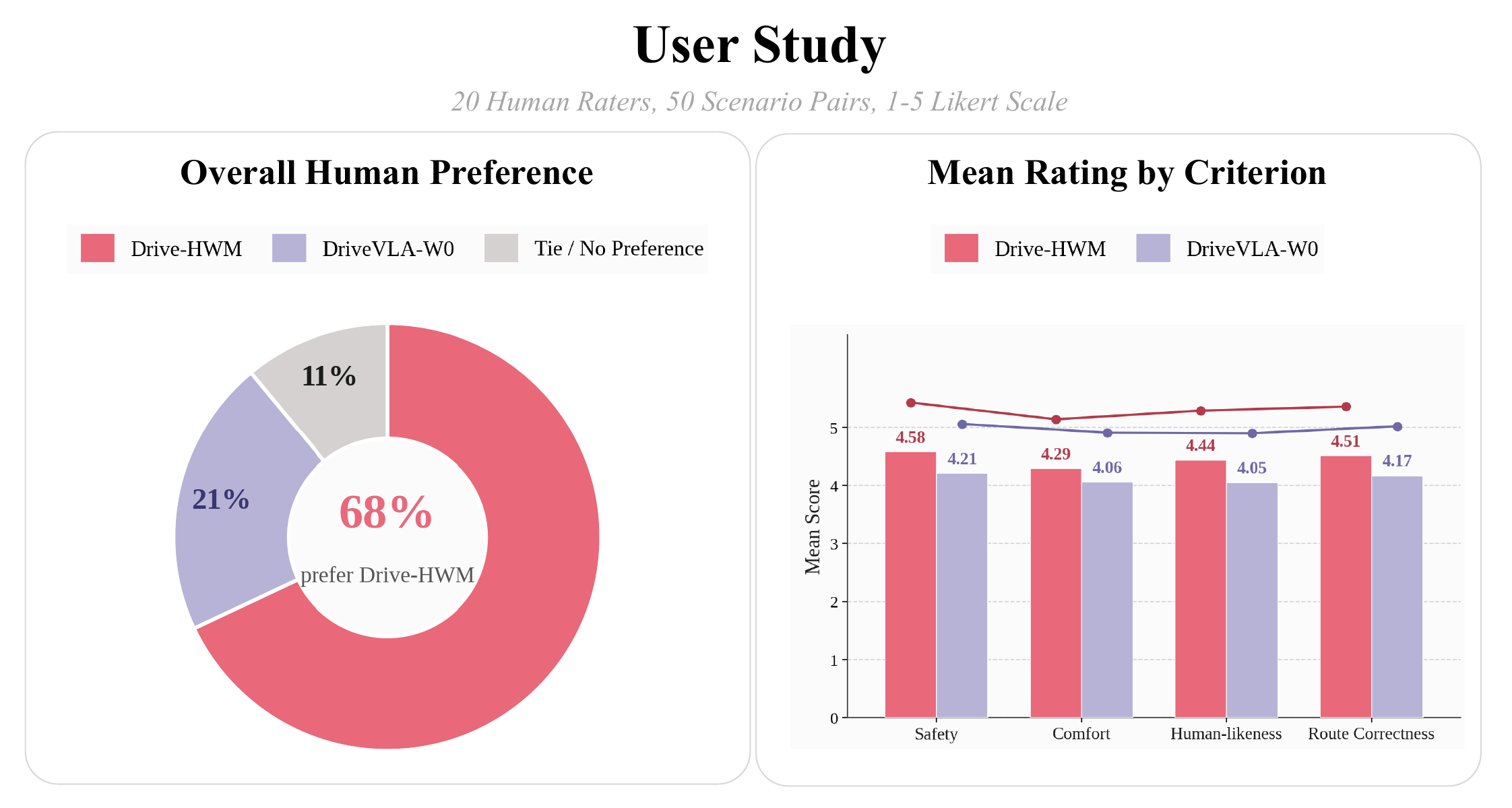}
    \caption{
    \textbf{User study of Drive-HWM and DriveVLA-W0.}
    Twenty human raters evaluate 50 paired driving scenarios.
    Drive-HWM is preferred in 68\% of comparisons and consistently receives higher ratings in safety, comfort, human-likeness, and route correctness.
    }
    \label{fig:user_study}
    \vspace{-0.2in}
\end{figure}

\paragraph{\textbf{User Study}}:
We further conduct a user study with 20 human raters over 50 paired driving scenarios to evaluate the perceived quality of the generated driving behaviors.
Participants compare Drive-HWM with DriveVLA-W0 and rate their behaviors on a 1--5 Likert scale in terms of safety, comfort, human-likeness, and route correctness.
As shown in Fig.~\ref{fig:user_study}, Drive-HWM is preferred in 68\% of the comparisons, compared with 21\% for DriveVLA-W0, while 11\% receive no clear preference.
Drive-HWM also achieves consistently higher mean ratings, with 4.58 vs.~4.21 in safety, 4.29 vs.~4.06 in comfort, 4.44 vs.~4.05 in human-likeness, and 4.51 vs.~4.17 in route correctness.
These results indicate that the advantages of Drive-HWM are also reflected in human perception, particularly in safety, human-like behavior, and route-level decision quality.

\section{Conclusion}
In this work, we investigated how future representation prediction and action
generation should be organized within a driving world model. Although these
two processes are closely related, they operate at different temporal scales:
future representations should capture coherent scene evolution over an
extended horizon, whereas executable actions must remain grounded in the
latest observation and be updated at a high frequency. To reconcile these
requirements, we proposed {Drive-HWM}, a hierarchical slow--fast world
modeling framework that couples multi-step future representation prediction
with immediate, observation-grounded action generation. The slow world model
learns {Dynamic-Aware Latents} through optical-flow supervision to
preserve the motion and interactions of the ego vehicle and surrounding
agents. Conditioned on these predictive representations and the latest
observation, the fast model jointly predicts the next RGB frame and the
immediate action, maintaining comprehensive scene information while allowing
each decision to be revised as new observations arrive. Extensive experiments
on NAVSIM v1 and v2 demonstrate the effectiveness of the proposed framework,
while comprehensive ablations validate the contributions of dynamics-aware
future prediction, next-frame visual supervision, and hierarchical
slow-to-fast conditioning. 

The existing version also exhibits limitations. First, the hierarchical slow--fast architecture introduces additional computational and memory overhead, motivating future research on model compression and asynchronous inference. Second, the current model does not explicitly capture multimodal futures or predictive uncertainty, which may limit its performance in ambiguous and rare driving scenarios.  We leave solving them as future works.

\section{Acknowledgment}
This work was supported by the New Generation Artificial Intelligence-National Science and Technology Major Project (2025ZD0122603). It was also supported by the Postdoctoral Fellowship Program and China Postdoctoral Science Foundation under Grant No. BX20250485, the Beijing Natural Science Foundation under Grant No. 4254100, by the Fundamental Research Funds for the Central Universities (JKF-2026071097445), and by Beijing Advanced Innovation Center for Future Blockchain and Privacy Computing.

% \begin{thebibliography}{1}
\bibliographystyle{IEEEtran}
\bibliography{main}

@String(ICCV  = {Int. Conf. Comput. Vis.})

@String(AAAI  = {AAAI})

@String(ICCV  = {ICCV})

@article{1,
  title={Enhancing end-to-end autonomous driving with latent world model},
  author={Li, Yingyan and Fan, Lue and He, Jiawei and Wang, Yuqi and Chen, Yuntao and Zhang, Zhaoxiang and Tan, Tieniu},
  journal={arXiv preprint arXiv:2406.08481},
  year={2024}
}

@inproceedings{wote,
  title={End-to-end driving with online trajectory evaluation via bev world model},
  author={Li, Yingyan and Wang, Yuqi and Liu, Yang and He, Jiawei and Fan, Lue and Zhang, Zhaoxiang},
  booktitle={Proceedings of the IEEE/CVF International Conference on Computer Vision},
  pages={27137--27146},
  year={2025}
}

@article{univla,
  title={Unified vision-language-action model},
  author={Wang, Yuqi and Li, Xinghang and Wang, Wenxuan and Zhang, Junbo and Li, Yingyan and Chen, Yuntao and Wang, Xinlong and Zhang, Zhaoxiang},
  journal={arXiv preprint arXiv:2506.19850},
  year={2025}
}

@article{worldvla,
  title={Worldvla: Towards autoregressive action world model},
  author={Cen, Jun and Yu, Chaohui and Yuan, Hangjie and Jiang, Yuming and Huang, Siteng and Guo, Jiayan and Li, Xin and Song, Yibing and Luo, Hao and Wang, Fan and others},
  journal={arXiv preprint arXiv:2506.21539},
  year={2025}
}

@article{autovla,
  title={Autovla: A vision-language-action model for end-to-end autonomous driving with adaptive reasoning and reinforcement fine-tuning},
  author={Zhou, Zewei and Cai, Tianhui and Zhao, Seth Z and Zhang, Yun and Huang, Zhiyu and Zhou, Bolei and Ma, Jiaqi},
  journal={arXiv preprint arXiv:2506.13757},
  year={2025}
}

@inproceedings{8,
  title={Bevformer v2: Adapting modern image backbones to bird's-eye-view recognition via perspective supervision},
  author={Yang, Chenyu and Chen, Yuntao and Tian, Hao and Tao, Chenxin and Zhu, Xizhou and Zhang, Zhaoxiang and Huang, Gao and Li, Hongyang and Qiao, Yu and Lu, Lewei and others},
  booktitle={Proceedings of the IEEE/CVF conference on computer vision and pattern recognition},
  pages={17830--17839},
  year={2023}
}

@article{9,
  title={Maptr: Structured modeling and learning for online vectorized hd map construction},
  author={Liao, Bencheng and Chen, Shaoyu and Wang, Xinggang and Cheng, Tianheng and Zhang, Qian and Liu, Wenyu and Huang, Chang},
  journal={arXiv preprint arXiv:2208.14437},
  year={2022}
}

@inproceedings{diffusiondrive,
  title={Diffusiondrive: Truncated diffusion model for end-to-end autonomous driving},
  author={Liao, Bencheng and Chen, Shaoyu and Yin, Haoran and Jiang, Bo and Wang, Cheng and Yan, Sixu and Zhang, Xinbang and Li, Xiangyu and Zhang, Ying and Zhang, Qian and others},
  booktitle={Proceedings of the Computer Vision and Pattern Recognition Conference},
  pages={12037--12047},
  year={2025}
}

@inproceedings{vad,
  title={Vad: Vectorized scene representation for efficient autonomous driving},
  author={Jiang, Bo and Chen, Shaoyu and Xu, Qing and Liao, Bencheng and Chen, Jiajie and Zhou, Helong and Zhang, Qian and Liu, Wenyu and Huang, Chang and Wang, Xinggang},
  booktitle={Proceedings of the IEEE/CVF International Conference on Computer Vision},
  pages={8340--8350},
  year={2023}
}

@inproceedings{drivedreamer,
  title={Drivedreamer: Towards real-world-drive world models for autonomous driving},
  author={Wang, Xiaofeng and Zhu, Zheng and Huang, Guan and Chen, Xinze and Zhu, Jiagang and Lu, Jiwen},
  booktitle={European conference on computer vision},
  pages={55--72},
  year={2024},
  organization={Springer}
}

@inproceedings{driveworld,
  title={Driveworld: 4d pre-trained scene understanding via world models for autonomous driving},
  author={Min, Chen and Zhao, Dawei and Xiao, Liang and Zhao, Jian and Xu, Xinli and Zhu, Zheng and Jin, Lei and Li, Jianshu and Guo, Yulan and Xing, Junliang and others},
  booktitle={Proceedings of the IEEE/CVF conference on computer vision and pattern recognition},
  pages={15522--15533},
  year={2024}
}

@article{gaia,
  title={Gaia-1: a generative world model for autonomous driving (2023)},
  author={Hu, Anthony and Russell, Lloyd and Yeo, Hudson and Murez, Zak and Fedoseev, George and Kendall, Alex and Shotton, Jamie and Corrado, Gianluca},
  journal={URL https://arxiv. org/abs/2309.17080},
  volume={3}
}

@article{vista,
  title={Vista: A generalizable driving world model with high fidelity and versatile controllability},
  author={Gao, Shenyuan and Yang, Jiazhi and Chen, Li and Chitta, Kashyap and Qiu, Yihang and Geiger, Andreas and Zhang, Jun and Li, Hongyang},
  journal={Advances in Neural Information Processing Systems},
  volume={37},
  pages={91560--91596},
  year={2024}
}

@inproceedings{uniad,
  title={Planning-oriented autonomous driving},
  author={Hu, Yihan and Yang, Jiazhi and Chen, Li and Li, Keyu and Sima, Chonghao and Zhu, Xizhou and Chai, Siqi and Du, Senyao and Lin, Tianwei and Wang, Wenhai and others},
  booktitle={Proceedings of the IEEE/CVF conference on computer vision and pattern recognition},
  pages={17853--17862},
  year={2023}
}

@article{11,
  title={Drivegpt4: Interpretable end-to-end autonomous driving via large language model},
  author={Xu, Zhenhua and Zhang, Yujia and Xie, Enze and Zhao, Zhen and Guo, Yong and Wong, Kwan-Yee K and Li, Zhenguo and Zhao, Hengshuang},
  journal={IEEE Robotics and Automation Letters},
  volume={9},
  number={10},
  pages={8186--8193},
  year={2024},
  publisher={IEEE}
}

@article{12,
  title={Ts-vlm: Text-guided softsort pooling for vision-language models in multi-view driving reasoning},
  author={Chen, Lihong and Hassani, Hossein and Nikan, Soodeh},
  journal={arXiv preprint arXiv:2505.12670},
  year={2025}
}

@article{13,
  title={DynRsl-VLM: Enhancing autonomous driving perception with dynamic resolution vision-language models},
  author={Zhou, Xirui and Shan, Lianlei and Gui, Xiaolin},
  journal={arXiv preprint arXiv:2503.11265},
  year={2025}
}

@article{14,
  title={Opendrivevla: Towards end-to-end autonomous driving with large vision language action model, 2025a},
  author={Zhou, Xingcheng and Han, Xuyuan and Yang, Feng and Ma, Yunpu and Knoll, Alois C},
  journal={URL https://arxiv. org/abs/2503.23463}
}

@article{15,
  title={Rag-driver: Generalisable driving explanations with retrieval-augmented in-context learning in multi-modal large language model},
  author={Yuan, Jianhao and Sun, Shuyang and Omeiza, Daniel and Zhao, Bo and Newman, Paul and Kunze, Lars and Gadd, Matthew},
  journal={arXiv preprint arXiv:2402.10828},
  year={2024}
}

@article{16,
  title={Safeauto: Knowledge-enhanced safe autonomous driving with multimodal foundation models},
  author={Zhang, Jiawei and Yang, Xuan and Wang, Taiqi and Yao, Yu and Petiushko, Aleksandr and Li, Bo},
  journal={arXiv preprint arXiv:2503.00211},
  year={2025}
}

@article{17,
  title={Covla: Comprehensive vision-language-action dataset for autonomous driving. In 2025 IEEE/CVF Winter Conference on Applications of Computer Vision (WACV)},
  author={Arai, Hidehisa and Miwa, Keita and Sasaki, Kento and Watanabe, Kohei and Yamaguchi, Yu and Aoki, Shunsuke and Yamamoto, Issei},
  journal={IEEE},
  volume={1},
  number={3},
  pages={4},
  year={2025}
}

@inproceedings{18,
  title={Simlingo: Vision-only closed-loop autonomous driving with language-action alignment},
  author={Renz, Katrin and Chen, Long and Arani, Elahe and Sinavski, Oleg},
  booktitle={Proceedings of the Computer Vision and Pattern Recognition Conference},
  pages={11993--12003},
  year={2025}
}

@article{19,
  title={CarLLaVA: Vision language models for camera-only closed-loop driving. arXiv 2024},
  author={Renz, K and Chen, L and Marcu, AM and H{\"u}nermann, J and Hanotte, B and Karnsund, A and Shotton, J and Arani, E and Sinavski, O},
  journal={arXiv preprint arXiv:2406.10165}
}

@article{diffvla,
  title={Diffvla: Vision-language guided diffusion planning for autonomous driving},
  author={Jiang, Anqing and Gao, Yu and Sun, Zhigang and Wang, Yiru and Wang, Jijun and Chai, Jinghao and Cao, Qian and Heng, Yuweng and Jiang, Hao and Dong, Yunda and others},
  journal={arXiv preprint arXiv:2505.19381},
  year={2025}
}

@article{20,
  title={Emma: End-to-end multimodal model for autonomous driving},
  author={Hwang, Jyh-Jing and Xu, Runsheng and Lin, Hubert and Hung, Wei-Chih and Ji, Jingwei and Choi, Kristy and Huang, Di and He, Tong and Covington, Paul and Sapp, Benjamin and others},
  journal={arXiv preprint arXiv:2410.23262},
  year={2024}
}

@article{drivemoe,
  title={DriveMoE: Mixture-of-experts for vision-language-action model in end-to-end autonomous driving},
  author={Yang, Zhenjie and Chai, Yilin and Jia, Xiaosong and Li, Qifeng and Shao, Yuqian and Zhu, Xuekai and Su, Haisheng and Yan, Junchi},
  journal={arXiv preprint arXiv:2505.16278},
  year={2025}
}

@article{21,
  title={Futuresightdrive: Thinking visually with spatio-temporal cot for autonomous driving},
  author={Zeng, Shuang and Chang, Xinyuan and Xie, Mengwei and Liu, Xinran and Bai, Yifan and Pan, Zheng and Xu, Mu and Wei, Xing and Guo, Ning},
  journal={arXiv preprint arXiv:2505.17685},
  year={2025}
}

@article{recogdrive,
  title={Recogdrive: A reinforced cognitive framework for end-to-end autonomous driving},
  author={Li, Yongkang and Xiong, Kaixin and Guo, Xiangyu and Li, Fang and Yan, Sixu and Xu, Gangwei and Zhou, Lijun and Chen, Long and Sun, Haiyang and Wang, Bing and others},
  journal={arXiv preprint arXiv:2506.08052},
  year={2025}
}

@misc{col4d,
  title={Learning unsupervised world models for autonomous driving via discrete diffusion},
  author={ZHANG, Lunjun and Xiong, Yuwen and Yang, Ze and ROMERO, Sergio CASAS and Urtasun, Raquel},
  year={2025},
  month=mar # "~27",
  publisher={Google Patents},
  note={US Patent App. 18/900,601}
}

@article{law,
  title={Enhancing end-to-end autonomous driving with latent world model},
  author={Li, Yingyan and Fan, Lue and He, Jiawei and Wang, Yuqi and Chen, Yuntao and Zhang, Zhaoxiang and Tan, Tieniu},
  journal={arXiv preprint arXiv:2406.08481},
  year={2024}
}

@article{emu3,
  title={Emu3: Next-token prediction is all you need},
  author={Wang, Xinlong and Zhang, Xiaosong and Luo, Zhengxiong and Sun, Quan and Cui, Yufeng and Wang, Jinsheng and Zhang, Fan and Wang, Yueze and Li, Zhen and Yu, Qiying and others},
  journal={arXiv preprint arXiv:2409.18869},
  year={2024}
}

@article{vjepa,
  title={Vl-jepa: Joint embedding predictive architecture for vision-language},
  author={Chen, Delong and Shukor, Mustafa and Moutakanni, Theo and Chung, Willy and Yu, Jade and Kasarla, Tejaswi and Bang, Yejin and Bolourchi, Allen and LeCun, Yann and Fung, Pascale},
  journal={arXiv preprint arXiv:2512.10942},
  year={2025}
}

@article{fast,
  title={Fast: Efficient action tokenization for vision-language-action models},
  author={Pertsch, Karl and Stachowicz, Kyle and Ichter, Brian and Driess, Danny and Nair, Suraj and Vuong, Quan and Mees, Oier and Finn, Chelsea and Levine, Sergey},
  journal={arXiv preprint arXiv:2501.09747},
  year={2025}
}

@article{navsim,
  title={Navsim: Data-driven non-reactive autonomous vehicle simulation and benchmarking},
  author={Dauner, Daniel and Hallgarten, Marcel and Li, Tianyu and Weng, Xinshuo and Huang, Zhiyu and Yang, Zetong and Li, Hongyang and Gilitschenski, Igor and Ivanovic, Boris and Pavone, Marco and others},
  journal={Advances in Neural Information Processing Systems},
  volume={37},
  pages={28706--28719},
  year={2024}
}

@article{nav2,
  title={Pseudo-simulation for autonomous driving},
  author={Cao, Wei and Hallgarten, Marcel and Li, Tianyu and Dauner, Daniel and Gu, Xunjiang and Wang, Caojun and Miron, Yakov and Aiello, Marco and Li, Hongyang and Gilitschenski, Igor and others},
  journal={arXiv preprint arXiv:2506.04218},
  year={2025}
}

@inproceedings{openscene,
  title={Openscene: The largest up-to-date 3d occupancy prediction benchmark in autonomous driving},
  author={Contributors, OpenScene},
  booktitle={Proceedings of the Conference on Computer Vision and Pattern Recognition, Vancouver, Canada},
  pages={18--22},
  year={2023}
}

@article{hydra,
  title={Hydra-mdp: End-to-end multimodal planning with multi-target hydra-distillation},
  author={Li, Zhenxin and Li, Kailin and Wang, Shihao and Lan, Shiyi and Yu, Zhiding and Ji, Yishen and Li, Zhiqi and Zhu, Ziyue and Kautz, Jan and Wu, Zuxuan and others},
  journal={arXiv preprint arXiv:2406.06978},
  year={2024}
}

@inproceedings{trans,
  title={Multi-modal fusion transformer for end-to-end autonomous driving},
  author={Prakash, Aditya and Chitta, Kashyap and Geiger, Andreas},
  booktitle={Proceedings of the IEEE/CVF conference on computer vision and pattern recognition},
  pages={7077--7087},
  year={2021}
}

@inproceedings{pdrive,
  title={Para-drive: Parallelized architecture for real-time autonomous driving},
  author={Weng, Xinshuo and Ivanovic, Boris and Wang, Yan and Wang, Yue and Pavone, Marco},
  booktitle={Proceedings of the IEEE/CVF Conference on Computer Vision and Pattern Recognition},
  pages={15449--15458},
  year={2024}
}

@article{drivevla,
  title={DriveVLA-W0: World models amplify data scaling law in autonomous driving},
  author={Li, Yingyan and Shang, Shuyao and Liu, Weisong and Zhan, Bing and Wang, Haochen and Wang, Yuqi and Chen, Yuntao and Wang, Xiaoman and An, Yasong and Tang, Chufeng and others},
  journal={arXiv preprint arXiv:2510.12796},
  year={2025}
}

@article{drivesuprim,
  title={Drivesuprim: Towards precise trajectory selection for end-to-end planning},
  author={Yao, Wenhao and Li, Zhenxin and Lan, Shiyi and Wang, Zi and Sun, Xinglong and Alvarez, Jose M and Wu, Zuxuan},
  journal={arXiv preprint arXiv:2506.06659},
  year={2025}
}

@article{artemis,
  title={Artemis: Autoregressive end-to-end trajectory planning with mixture of experts for autonomous driving},
  author={Feng, Renju and Xi, Ning and Chu, Duanfeng and Wang, Rukang and Deng, Zejian and Wang, Anzheng and Lu, Liping and Wang, Jinxiang and Huang, Yanjun},
  journal={IEEE Robotics and Automation Letters},
  volume={11},
  number={1},
  pages={226--233},
  year={2025},
  publisher={IEEE}
}

@article{nuplan,
  title={nuplan: A closed-loop ml-based planning benchmark for autonomous vehicles},
  author={Caesar, Holger and Kabzan, Juraj and Tan, Kok Seang and Fong, Whye Kit and Wolff, Eric and Lang, Alex and Fletcher, Luke and Beijbom, Oscar and Omari, Sammy},
  journal={arXiv preprint arXiv:2106.11810},
  year={2021}
}

@article{cogvideo,
  title={Cogvideo: Large-scale pretraining for text-to-video generation via transformers},
  author={Hong, Wenyi and Ding, Ming and Zheng, Wendi and Liu, Xinghan and Tang, Jie},
  journal={arXiv preprint arXiv:2205.15868},
  year={2022}
}

@article{wan,
  title={Wan: Open and advanced large-scale video generative models},
  author={Wan, Team and Wang, Ang and Ai, Baole and Wen, Bin and Mao, Chaojie and Xie, Chen-Wei and Chen, Di and Yu, Feiwu and Zhao, Haiming and Yang, Jianxiao and others},
  journal={arXiv preprint arXiv:2503.20314},
  year={2025}
}

@inproceedings{film,
  title={Film: Visual reasoning with a general conditioning layer},
  author={Perez, Ethan and Strub, Florian and De Vries, Harm and Dumoulin, Vincent and Courville, Aaron},
  booktitle={Proceedings of the AAAI conference on artificial intelligence},
  volume={32},
  number={1},
  year={2018}
}

@article{qian2024fasionad,
  title={Fasionad: Fast and slow fusion thinking systems for human-like autonomous driving with adaptive feedback},
  author={Qian, Kangan and Ma, Zhikun and He, Yangfan and Luo, Ziang and Shi, Tianyu and Zhu, Tianze and Li, Jiayin and Wang, Jianhui and Chen, Ziyu and He, Xiao and others},
  journal={arXiv preprint arXiv:2411.18013},
  year={2024}
}

@article{luo2025adathinkdrive,
  title={Adathinkdrive: Adaptive thinking via reinforcement learning for autonomous driving},
  author={Luo, Yuechen and Li, Fang and Xu, Shaoqing and Lai, Zhiyi and Yang, Lei and Chen, Qimao and Luo, Ziang and Xie, Zixun and Jiang, Shengyin and Liu, Jiaxin and others},
  journal={arXiv preprint arXiv:2509.13769},
  year={2025}
}

@article{awais2025foundation,
  title={Foundation models defining a new era in vision: a survey and outlook},
  author={Awais, Muhammad and Naseer, Muzammal and Khan, Salman and Anwer, Rao Muhammad and Cholakkal, Hisham and Shah, Mubarak and Yang, Ming-Hsuan and Khan, Fahad Shahbaz},
  journal={IEEE Transactions on Pattern Analysis and Machine Intelligence},
  volume={47},
  number={4},
  pages={2245--2264},
  year={2025},
  publisher={IEEE}
}

@article{lin2025navcot,
  title={Navcot: Boosting llm-based vision-and-language navigation via learning disentangled reasoning},
  author={Lin, Bingqian and Nie, Yunshuang and Wei, Ziming and Chen, Jiaqi and Ma, Shikui and Han, Jianhua and Xu, Hang and Chang, Xiaojun and Liang, Xiaodan},
  journal={IEEE Transactions on Pattern Analysis and Machine Intelligence},
  volume={47},
  number={7},
  pages={5945--5957},
  year={2025},
  publisher={IEEE}
}

@article{wang2024jarvis,
  title={Jarvis-1: Open-world multi-task agents with memory-augmented multimodal language models},
  author={Wang, Zihao and Cai, Shaofei and Liu, Anji and Jin, Yonggang and Hou, Jinbing and Zhang, Bowei and Lin, Haowei and He, Zhaofeng and Zheng, Zilong and Yang, Yaodong and others},
  journal={IEEE Transactions on Pattern Analysis and Machine Intelligence},
  volume={47},
  number={3},
  pages={1894--1907},
  year={2024},
  publisher={IEEE}
}

@inproceedings{zeng2026rethinking,
  title={Rethinking driving world model as synthetic data generator for perception tasks},
  author={Zeng, Kai and Wu, Zhanqian and Xiong, Kaixin and Wei, Xiaobao and Guo, Xiangyu and Zhu, Zhenxin and Ho, Kalok and Zhou, Lijun and Zeng, Bohan and Lu, Ming and others},
  booktitle={International Conference on Learning Representations},
  volume={2026},
  pages={135507--135534},
  year={2026}
}

@inproceedings{he2026pre,
  title={Pre-trained video generative models as world simulators},
  author={He, Haoran and Zhang, Yang and Lin, Liang and Xu, Zhongwen and Pan, Ling},
  booktitle={Proceedings of the AAAI Conference on Artificial Intelligence},
  volume={40},
  number={6},
  pages={4645--4653},
  year={2026}
}

@inproceedings{zhang2026world,
  title={World-in-world: World models in a closed-loop world},
  author={Zhang, Jiahan and Jiang, Muqing and Dai, Nanru and Lu, Taiming and Uzunoglu, Arda and Zhang, Shunchi and Wei, Yana and Wang, Jiahao and Patel, Vishal and Liang, Paul and others},
  booktitle={International Conference on Learning Representations},
  volume={2026},
  pages={55660--55699},
  year={2026}
}

@article{jiang2025irl,
  title={Irl-vla: Training an vision-language-action policy via reward world model},
  author={Jiang, Anqing and Gao, Yu and Wang, Yiru and Sun, Zhigang and Wang, Shuo and Heng, Yuwen and Sun, Hao and Tang, Shichen and Zhu, Lijuan and Chai, Jinhao and others},
  journal={arXiv preprint arXiv:2508.06571},
  year={2025}
}

@article{ye2026gigaworld,
  title={GigaWorld-Policy: An Efficient Action-Centered World--Action Model},
  author={Ye, Angen and Wang, Boyuan and Ni, Chaojun and Huang, Guan and Zhao, Guosheng and Li, Hao and Li, Hengtao and Li, Jie and Lv, Jindi and Liu, Jingyu and others},
  journal={arXiv preprint arXiv:2603.17240},
  year={2026}
}

@inproceedings{bi2026motus,
  title={Motus: A unified latent action world model},
  author={Bi, Hongzhe and Tan, Hengkai and Xie, Shenghao and Wang, Zeyuan and Huang, Shuhe and Liu, Haitian and Zhao, Ruowen and Feng, Yao and Xiang, Chendong and Rong, Yinze and others},
  booktitle={Proceedings of the IEEE/CVF Conference on Computer Vision and Pattern Recognition},
  pages={35101--35113},
  year={2026}
}

@inproceedings{zhu2026wmpo,
  title={Wmpo: World model-based policy optimization for vision-language-action models},
  author={Zhu, Fangqi and Yan, Zhengyang and Hong, Zicong and Shou, Quanxin and Ma, Xiao and Guo, Song},
  booktitle={International Conference on Learning Representations},
  volume={2026},
  pages={62486--62502},
  year={2026}
}

@inproceedings{li2025end,
  title={End-to-end driving with online trajectory evaluation via bev world model},
  author={Li, Yingyan and Wang, Yuqi and Liu, Yang and He, Jiawei and Fan, Lue and Zhang, Zhaoxiang},
  booktitle={2025 IEEE/CVF International Conference on Computer Vision (ICCV)},
  pages={27137--27146},
  year={2025},
  organization={IEEE}
}

@article{wang2025prophetdwm,
  title={Prophetdwm: A driving world model for rolling out future actions and videos},
  author={Wang, Xiaodong and Peng, Peixi},
  journal={arXiv preprint arXiv:2505.18650},
  year={2025}
}

@article{li2024llava,
  title={Llava-onevision: Easy visual task transfer},
  author={Li, Bo and Zhang, Yuanhan and Guo, Dong and Zhang, Renrui and Li, Feng and Zhang, Hao and Zhang, Kaichen and Zhang, Peiyuan and Li, Yanwei and Liu, Ziwei and others},
  journal={arXiv preprint arXiv:2408.03326},
  year={2024}
}

@article{bai2025qwen3,
  title={Qwen3-vl technical report},
  author={Bai, Shuai and Cai, Yuxuan and Chen, Ruizhe and Chen, Keqin and Chen, Xionghui and Cheng, Zesen and Deng, Lianghao and Ding, Wei and Gao, Chang and Ge, Chunjiang and others},
  journal={arXiv preprint arXiv:2511.21631},
  year={2025}
}

@article{guo2022adaln,
  title={Adaln: a vision transformer for multidomain learning and predisaster building information extraction from images},
  author={Guo, Yunhui and Wang, Chaofeng and Yu, Stella X and McKenna, Frank and Law, Kincho H},
  journal={Journal of Computing in Civil Engineering},
  volume={36},
  number={5},
  pages={04022024},
  year={2022},
  publisher={American Society of Civil Engineers}
}

@article{alayrac2022flamingo,
  title={Flamingo: a visual language model for few-shot learning},
  author={Alayrac, Jean-Baptiste and Donahue, Jeff and Luc, Pauline and Miech, Antoine and Barr, Iain and Hasson, Yana and Lenc, Karel and Mensch, Arthur and Millican, Katherine and Reynolds, Malcolm and others},
  journal={Advances in neural information processing systems},
  volume={35},
  pages={23716--23736},
  year={2022}
}

\vspace{-3em}
\begin{IEEEbiography}[{\includegraphics[width=1in,height=1.2in,clip,keepaspectratio]{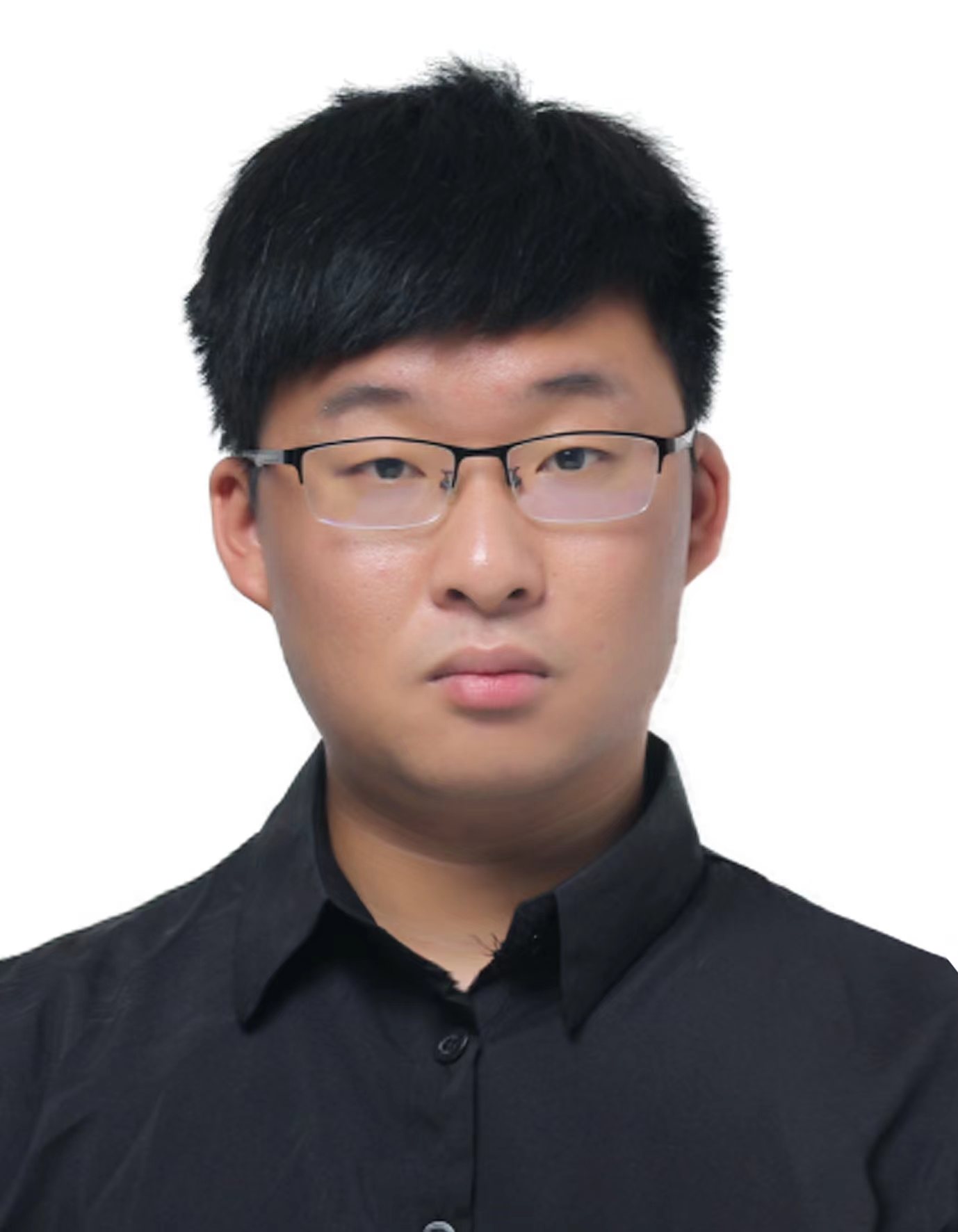}}]{Zhaoxin Fan} received his Ph.D. degree from the School of Information, Renmin University, China in 2024. He has also conducted research at Carnegie Mellon University and  Hong Kong University of Science and Technology. He is currently an Assistant Professor in the Institute of Artificial Intelligence, Beihang University. His research interests include multi-modal large language models , computer vision, and embodied AI.
\end{IEEEbiography}
\vspace{-3em}
\begin{IEEEbiography}[{\includegraphics[width=1in,height=1.2in,clip,keepaspectratio]{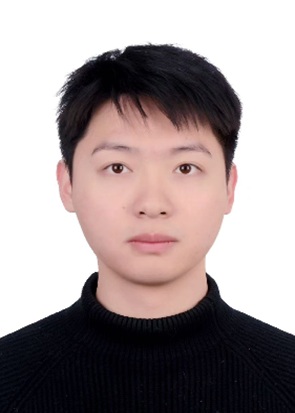}}]{Tianbao Zhang} is currently an PhD student at Shanghai Jiao Tong University (SJTU) and the Intern at Dim12 AI. He won CVPR 2026 GigaBrain Challenge 2026 Workshop (Best Paper Runner Up) and won Chinese Conference on Pattern Recognition and Computer Vision (PRCV Best Student Papera and CCF outstanding Paper). His research interests include computer vision, avatars, and embodied AI. 
\end{IEEEbiography}
\vspace{-3em}
\begin{IEEEbiography}[{\includegraphics[width=1in,height=1.2in,clip,keepaspectratio]{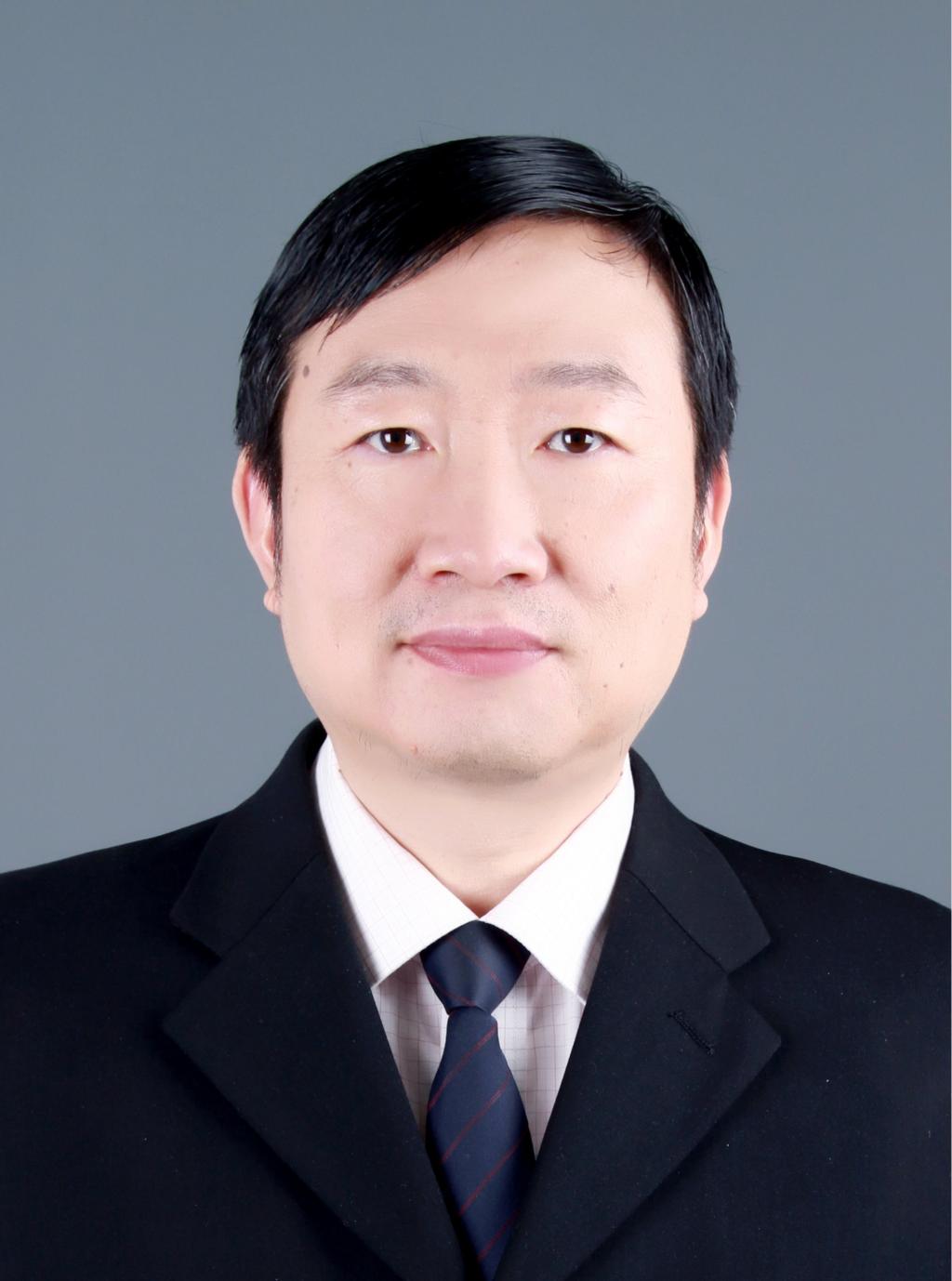}}]{Wenjun Wu} received the PhD degree in computer science from Beihang University, in 2001. He was
employed with Argonne National Laboratory as a
research scientist working on grid computing, cloud
computing, media collaboration, etc., until 2012. He
is currently employed with Beihang University as a
professor. His research interests include crowdsourcing, machine learning, cloud computing, eScience,
and cyber infrastructure.
\end{IEEEbiography}

\vspace{-3em}

\begin{IEEEbiography}[{\includegraphics[width=1in,height=1.2in,clip,keepaspectratio]{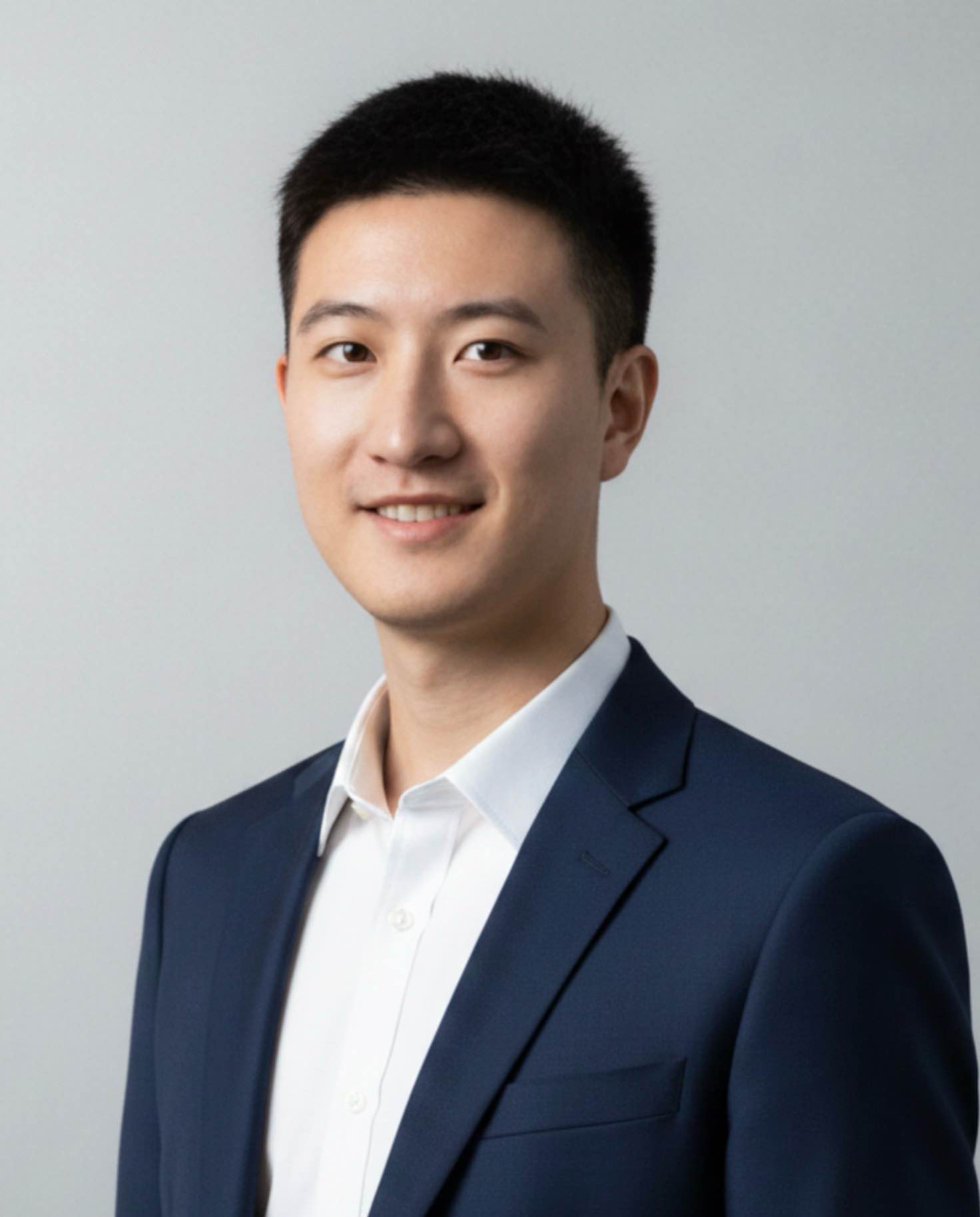}}]{Xiaofeng Wang} received the Ph.D. degree from the Institute of Automation, Chinese Academy of Sciences, Beijing, China, in 2025. His research interests include general world models. He is currently the Deputy Director of the Beijing Key Laboratory of General World Models and an Algorithm Partner at Excellent Vision Technology. 
\end{IEEEbiography}

\vspace{-2em}

\begin{IEEEbiography}[{\includegraphics[width=1in,height=1.2in,clip,keepaspectratio]{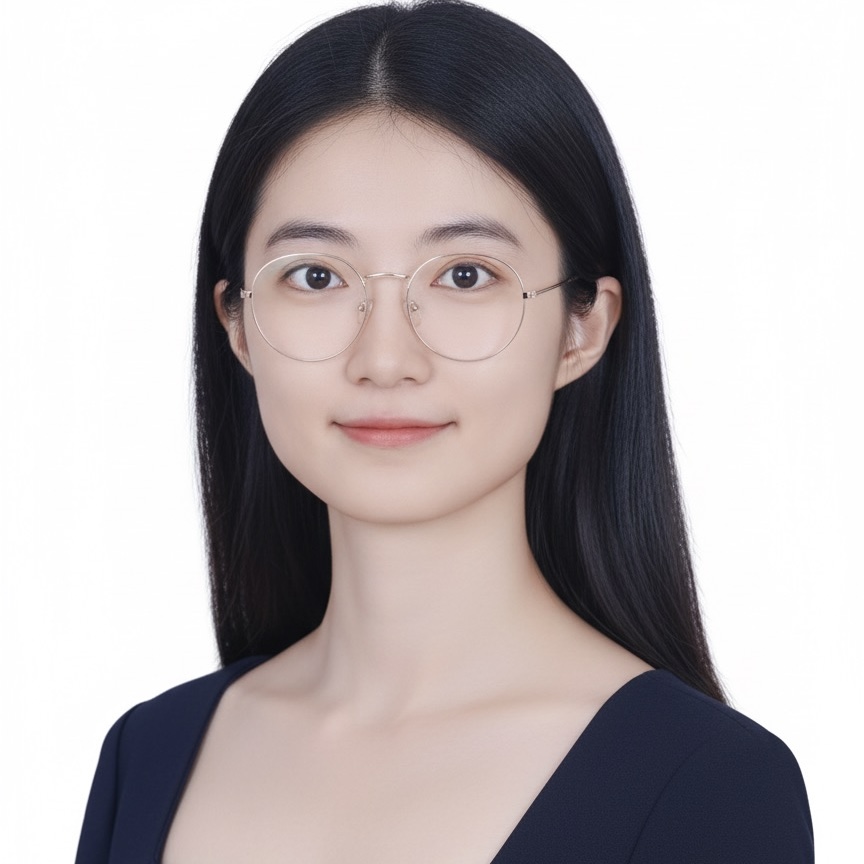}}]{Yeying Jin} currently a Staff Researcher with Tencent, Singapore, where she is the Research Lead for world models, agentic AI, and AIGC, and she is also an Adjunct Faculty Member with the Department of Electrical and Computer Engineering, NUS. Her current research interests include game world models, agentic video generation, and coding agents. She has authored or coauthored more than 70 papers at top-tier venues, including 12 as first author, which have received over 1900 citations, with more than 1100 citations for her first-author work.
\end{IEEEbiography}
\vspace{-2em}
\begin{IEEEbiography}[{\includegraphics[width=1in,height=1in,clip,keepaspectratio]{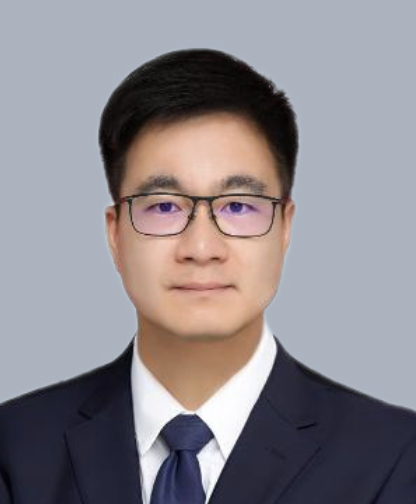}}]{Jian Zhao} is currently the Leader of EVOL Lab and a Principal Research Scientist with the Institute of AI (TeleAI), China Telecom, and a Researcher and Ph.D. Supervisor with School of Artificial Intelligence, Optics and Electronics
(iOPEN), Northwestern Polytechnical University (NWPU).
\end{IEEEbiography}

\vspace{-2em}
\begin{IEEEbiography}[{\includegraphics[width=1in,height=1.2in,clip,keepaspectratio]{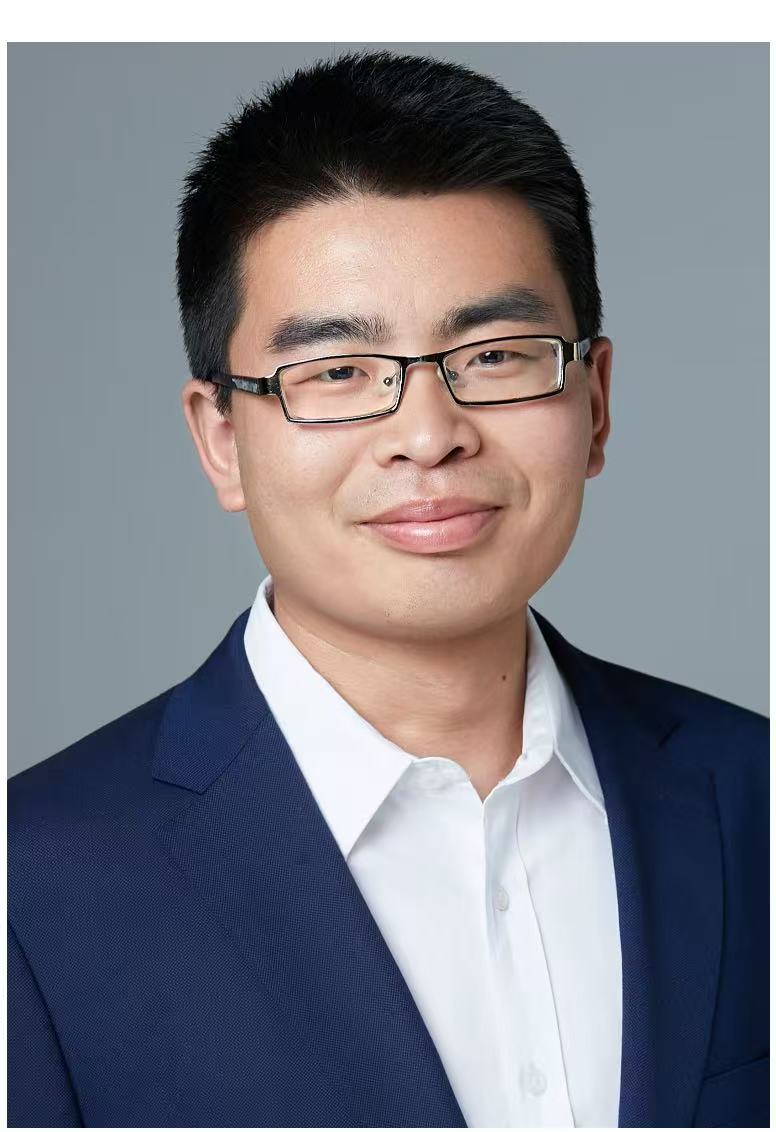}}]{Zheng Zhu} is currently the Co-founder and Chief Scientist at GigaAI. During 2019-2021, he was a post-doc fellow at Tsinghua University. Before that, he received Ph.D. degree from Institute of Automation, Chinese Academy of Sciences in 2019. During 2016-2019, he was research interns at SenseTime, Horizon Robotics, and DeepGlint. He has co-authored more than 80 top journal and conference papers mainly on computer vision and robotics problems.
\end{IEEEbiography}
\vspace{-2em}

\begin{IEEEbiography}[{\includegraphics[width=1in,height=1.2in,clip,keepaspectratio]{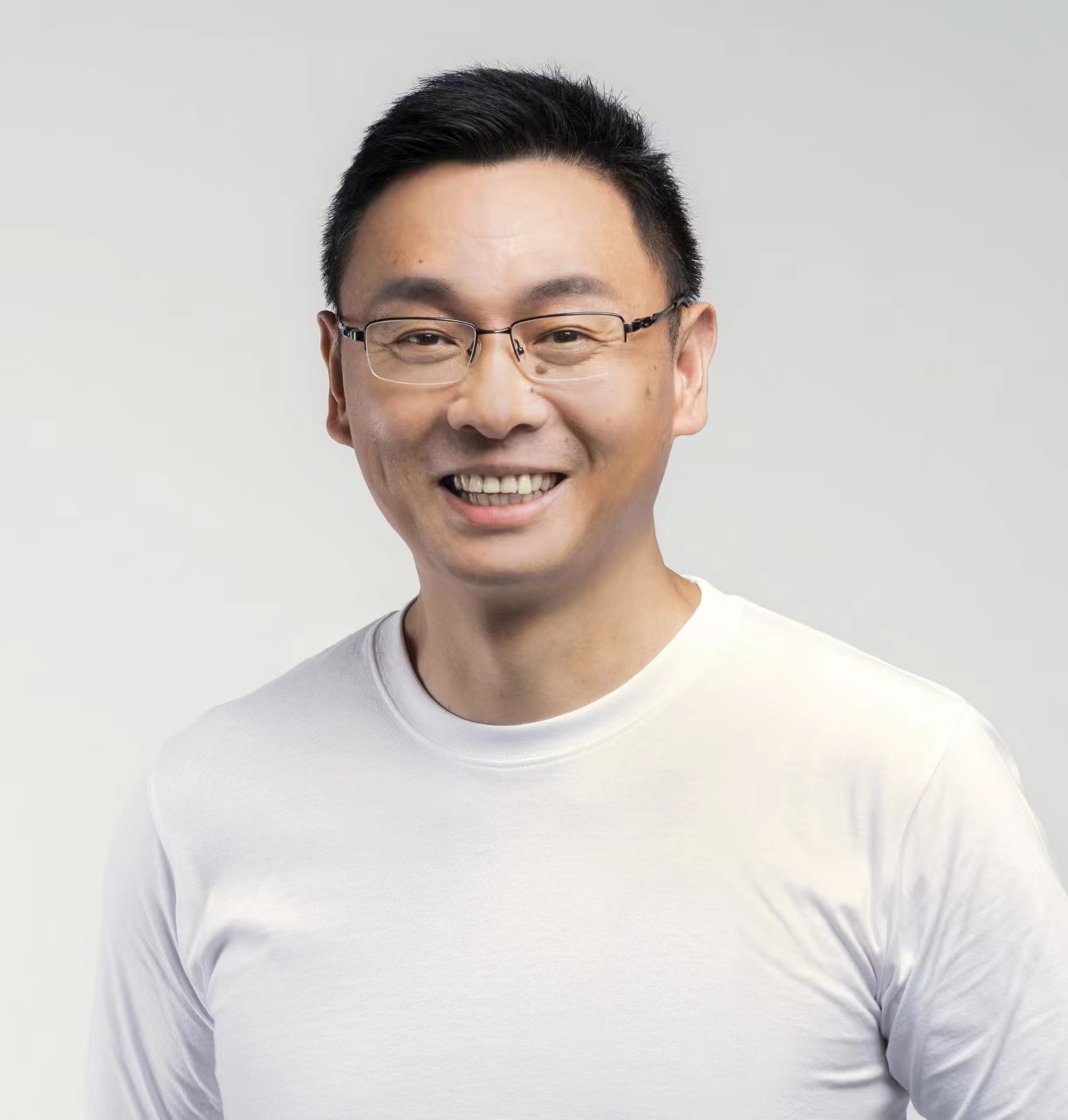}}]{Shuicheng Yan} is currently a Distinguished Professor (Practice) at the School of Computing, National University of Singapore. He previously served as the Group Chief Scientist at Sea Group and has held several other notable industry positions. Prof. Yan is a Fellow of the Singapore Academy of Engineering, AAAI, ACM, IEEE, and IAPR. His research focuses on computer vision, machine learning, and multimedia analysis. He has also been recognised as one of the World's Highly Cited Researchers for ten years.
\end{IEEEbiography}

\end{document}